\documentclass[conference]{IEEEtran}

\usepackage{graphicx}
\usepackage{amsmath}

\usepackage{amssymb}
\usepackage{booktabs}
\usepackage{url}
\let\labelindent\relax
\usepackage{hyperref}
\usepackage{epsfig}
\usepackage{enumitem}
\setlist[itemize]{leftmargin=3mm}
\include{_header}
\usepackage{multirow}
\usepackage{makecell}
\usepackage{xr}
\usepackage{enumitem}
\usepackage{tikz}
\usepackage[framemethod=tikz]{mdframed}

\ifCLASSINFOpdf
\else
\fi

\begin{document}
\title{Rethinking the Threat and Accessibility of Adversarial Attacks against Face \\Recognition Systems}

\author{\IEEEauthorblockN{Yuxin Cao\IEEEauthorrefmark{1},
Yumeng Zhu\IEEEauthorrefmark{2},
Derui Wang\IEEEauthorrefmark{3}, 
Sheng Wen\IEEEauthorrefmark{4}, 
Minhui Xue\IEEEauthorrefmark{3}, 
Jin Lu\IEEEauthorrefmark{2} and
Hao Ge\IEEEauthorrefmark{2}}
\IEEEauthorblockA{\IEEEauthorrefmark{1}National University of Singapore, Singapore}
\IEEEauthorblockA{\IEEEauthorrefmark{2}Ping An Technology, China}
\IEEEauthorblockA{\IEEEauthorrefmark{3}CSIRO’s Data61, Australia}
\IEEEauthorblockA{\IEEEauthorrefmark{4}Swinburne University of Technology, Australia}}

\maketitle

\begin{abstract}
Face recognition pipelines have been widely deployed in various mission-critical systems in trust, equitable and responsible AI applications. However, the emergence of adversarial attacks has threatened the security of the entire recognition pipeline. Despite the sheer number of attack methods proposed for crafting adversarial examples in both digital and physical forms, it is never an easy task to assess the real threat level of different attacks and obtain useful insight into the key risks confronted by face recognition systems. Traditional attacks view imperceptibility as the most important measurement to keep perturbations stealthy, while we suspect that industry professionals may possess a different opinion. In this paper, we delve into measuring the threat brought about by adversarial attacks from the perspectives of the industry and the applications of face recognition. In contrast to widely studied sophisticated attacks in the field, we propose an effective yet easy-to-launch physical adversarial attack, named AdvColor, against black-box face recognition pipelines in the physical world. AdvColor evades and fools models in the recognition pipeline via directly supplying printed photos of human faces to the system under adversarial illuminations. Experimental results show that physical AdvColor examples can achieve a fooling rate of more than 96\% against the anti-spoofing model and an overall attack success rate of 88\% against the face recognition pipeline. More importantly, we conduct a survey on the threats of prevailing adversarial attacks, including AdvColor, to understand the gap between the machine-measured and human-assessed threat levels of different forms of adversarial attacks. The survey results surprisingly indicate that, compared to deliberately launched imperceptible attacks, perceptible but accessible attacks pose more lethal threats to real-world commercial systems of face recognition. 
\end{abstract}

\section{Introduction}
Deep Neural Networks (DNNs) are known to be vulnerable to adversarial attacks~\cite{szegedy2014intriguing,goodfellow2015explaining}. Current attacks focus on crafting human-imperceptible adversarial examples by using either $\ell_p$-bounded (restricted) or semantic (unrestricted) perturbations to maintain the stealthiness of the attacks. As a constraint/objective, imperceptibility appears in almost all image attack methods, inherited from the very first literature of adversarial examples~\cite{szegedy2014intriguing}. Traditional $\ell_p$-bounded attacks~\cite{akhtar2018threat,qiu2019review,chakraborty2021survey} are challenged by the following three limitations. \textit{(i)} \textbf{High cost}: optimizing $\ell_p$ perturbations may require a large amount of computing and high-performance hardware support. \textit{(ii)} \textbf{Low efficiency}: a large number of queries (usually in the order of $10^4 \sim 10^6 $) are needed to succeed in black-box attacks. \textit{(iii)} \textbf{Limited physical operability}: it is difficult to instantiate perturbations through physical devices in the real world. Despite some semantic-based digital/physical attacks in the image domain~\cite{wang2019adversarial,shamsabadi2020colorfool,duan2020camouflage,lovisotto2021slap}, the limitations remain unsolved. For instance, in many human-unattended scenarios, such as access control systems, the perceptibility of the perturbations may no longer be crucial. Instead, easy-to-execute but effective attacks that can be launched at low cost may impose even severer security risks to the systems in the industry. Henceforth, a question to be asked is: 

 \begin{mdframed}[backgroundcolor=white!10,rightline=true,leftline=true,topline=true,bottomline=true,roundcorner=2mm,everyline=true,nobreak=false]  
\emph{Does imperceptibility really matter that much in the constitution of high-threat-level adversaries?} 
 \end{mdframed}
\vspace{-3mm}

\begin{figure}[t]
\begin{center}
  \includegraphics[width=0.75\linewidth]{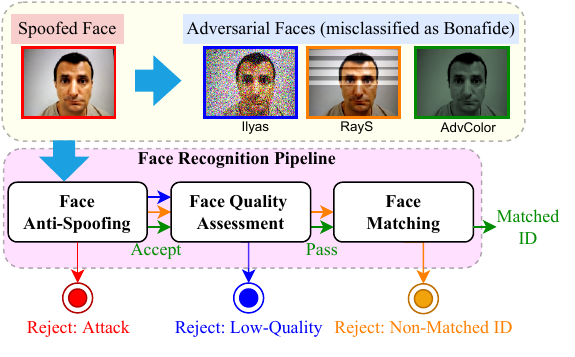}
\end{center}
\vspace{-3mm}
  \caption{The face recognition pipeline. The spoofed face is rejected in the face anti-spoofing stage, while the adversarial face generated by AdvColor passes the whole pipeline.
  }
\label{fig:pipeline}
\vspace{-2mm}
\end{figure}

In this paper, we specifically focus on the case of face recognition systems as they are widely deployed in trust, equitable and responsible AI applications and attract great concerns in their security, privacy and safety~\cite{xue2023rai4ioe,chen2024rethinking}. According to a recent statistical report~\cite{max202437+}, 131 million Americans use face recognition systems daily, with the most widespread applications including unlocking personal electronic devices, logging into apps, and accessing bank accounts. The use of face recognition systems to improve security at airports and banks has achieved an acceptance rate of over 54\% among Americans, and the use of face recognition systems in passport control and building access has received over 72\% approval. However, such quotidian techniques have been proven to be vulnerable in recent studies~\cite{vakhshiteh2021adversarial,li2023sibling}. The emergence of presentation attacks is strongly threatening such systems through the fabrication of spoofed face images (\eg print, replay, mask)~\cite{marcel2019handbook,ramachandra2017presentation,goswami2018unravelling}. Some researchers have shown that bypassing Apple’s FaceID is possible simply by wearing a pair of modified glasses~\cite{bypass_apple}. To distinguish between spoofed faces and bonafide ones, face anti-spoofing models, oftentimes together with a face quality assessment, are added to face recognition systems~\cite{li2018face,de2013can}. In other words, face recognition systems can be regarded as a three-stage authentication: face anti-spoofing classification, face quality assessment, and face matching~\cite{mohanty2021multi,yu2022deep}, as shown in Figure~\ref{fig:pipeline}. Nevertheless, most anti-spoofing models are based on DNNs. This has raised great concerns about the security of face recognition systems~\cite{dong2019efficient,sharif2016accessorize,zhang2020adversarial}. An attacker may use a victim's spoofed face to circumvent the face anti-spoofing model (\ie the classification model misclassifies the spoofed face as bonafide) while maintaining the face quality score and face ID matching results. This can incur severe aftermaths, and the victim may even be unaware of such attacks from beginning to end. Therefore, the robustness of face anti-spoofing/recognition models needs to be improved and well protected. 

\begin{figure}[t]
\begin{center}
  \includegraphics[width=.7\linewidth]{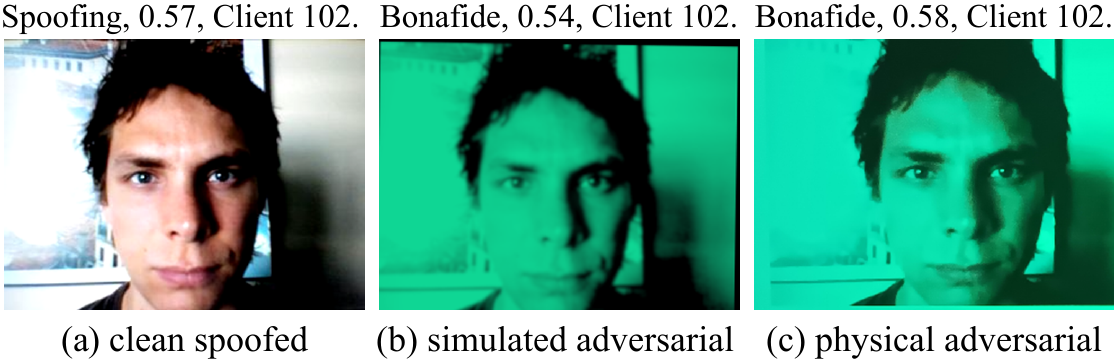}
\end{center}
\vspace{-3mm}
  \caption{Examples before/after the AdvColor attack. Texts above images represent the prediction of the anti-spoofing model, the face quality score, and the face matching result.
}
\label{fig:examples}
\vspace{-2mm}
\end{figure}

To unveil the answer to the question regarding imperceptibility, we first construct a strong but perceptible adversary through a black-box physical attack named \textit{\textbf{AdvColor}}. AdvColor fools face recognition systems by simply changing the environmental lighting in printed portraits with common illumination devices. AdvColor has the following advantages: 
\textit{(i)} AdvColor is easy-to-implement and only costs a little (\eg a smart light bulb). 
\textit{(ii)} Compared to attacks in the image domain, AdvColor can reduce the number of queries by several orders of magnitude and meanwhile achieve a high fooling rate. 
\textit{(iii)} AdvColor creates reusable specifications to produce adversarial illumination through ubiquitous devices in the real world.
We found that most parts of the face recognition pipeline are surprisingly vulnerable to the attack. In particular, once an AdvColor example bypasses the face anti-spoofing module in the pipeline under a satisfactory face quality score, it can effortlessly fool the rest of the system used for face matching, as shown in Figure~\ref{fig:examples}. In addition, AdvColor can provide several universally adversarial RGB values to which face recognition systems are vulnerable. These RGB values can help create universal adversarial illuminations that can be used by different attackers over time. Furthermore, the AdvColor examples can easily bypass a series of dominant defense methods. Subsequently, we conduct a survey to figure out the important characteristics of attacks that industry professionals are concerned about with greater threat. The survey shows evidence that, unlike the opinion in academia, the sophistication and imperceptibility of attacks are not a great windvane for high risks. Instead, human experience suggests that simple but easy-to-exploit attacks are extremely harmful and threatening, which impose a negative influence on business, customers, privacy leakage and other aspects. The source code of AdvColor is available at \url{https://github.com/AdvColor123/AdvColor}, together with main questions of our survey.

\noindent \textbf{Our main contributions are as follows.}
\begin{itemize}
    \item We propose a query-efficient black-box physical adversarial attack, named AdvColor, against face recognition systems. AdvColor is reusable, low-cost, and highly effective.

    \item We carry out a survey to explore the cognition gap between lab-measured and real-world threats.
    Results show that industrial insiders care more about perceptible but low-cost attacks rather than theoretically complicated but imperceptible attacks. 

    \item We propose new measurements as the replacement for imperceptibility towards quantifying the threat level
    of adversarial attacks. Industry professionals express that there are already simple but effective attacks similar to AdvColor that have threatened business products and may have far-reaching negative impacts/aftermaths in multiple aspects. They are appealing for academic attention to such attacks and potential mitigation.
\end{itemize}

\section{Adversaries in the Field}
To begin with, we take an overview of the adversarial attacks available to possible adversaries that aim to sabotage face recognition applications. We consider not only specialized attacks but also broad attacks in the image/video domain, since these attacks can be adapted against the face recognition task. 

\subsection{General Adversarial Attacks}\label{subsec:taxonomy}
We summarize the existing widely recognized adversarial attacks for images/videos and face recognition and separate them from the aspect of perturbation form. We divided them into restricted and unrestricted attacks. In restricted attacks, the adversary is merely allowed to alter to pixels within a range of $\ell_p$ norms, \ie $\left\| x_{adv} - x \right\|_p < \varepsilon$, where $x$ and $x_{adv}$ denote the original image and adversarial image, respectively, $\varepsilon$ stands for the perturbation threshold to guarantee the unobservability of human eyes. This kind of attack is regarded the cornerstone of adversarial attacks when they came to light and has gained widespread recognition in the past decade. A plethora of work was made public considering the
$\ell_\infty$~\cite{goodfellow2015explaining,kurakin2018adversarial,chatzikyriakidis2019adversarial}, $\ell_2$~\cite{szegedy2014intriguing,moosavi2016deepfool,jang2017objective,guo2019simple,dong2019efficient}, $\ell_0$~\cite{papernot2016limitations,su2019one} or $\ell_1$~\cite{chen2018ead} norm to restrict perturbations. Some methods with stronger generalization take multiple norm constraints simultaneously into account, including white-box attacks~\cite{moosavi2017universal,carlini2017towards,madry2017towards,hirano2020simple,ghiasi2019breaking}, where attackers have access to the model architecture and parameters, and black-box attacks~\cite{chen2017zoo,andriushchenko2020square,chen2020hopskipjumpattack,brendel2019accurate,cisse2017houdini,zhong2020towards}, where attackers have no knowledge of the surrogate model.

Another branch of attacks, \ie unrestricted attacks, cannot be overlooked and underestimated since perturbations do not have a crucial difference in the semantic information of the images, making adversarial examples simpler and practical, especially when deployed in physical scenarios. Dominant unrestricted attacks can be roughly divided into attacks based on patch~\cite{sharif2016accessorize,brown2017adversarial,wang2019advpattern,thys2019fooling,hu2021naturalistic,cheng2022physical,nesti2022evaluating,yang2023towards,xiao2021improving,pautov2019adversarial,komkov2021advhat}, watermark~\cite{wang2019adversarial,jia2020adv,jiang2021fawa,wu2021universal,zuo2023mispso}~\footnote{We exclude another branch of watermark attacks where the watermark are embedded in the target image (\eg ~\cite{feng2021digital}), since there is no natural visual effect; these attacks cannot be regarded as unrestricted attacks.}, optical~\cite{guo2020watch,lovisotto2021slap,gnanasambandam2021optical,duan2021adversarial,wang2023rfla,li2023physical,nguyen2020adversarial,zhou2018invisible}, style~\cite{duan2020camouflage,cao2023stylefool}, texture~\cite{hu2022adversarial,hu2023physically} and color~\cite{shamsabadi2020colorfool,yuan2022natural,hu2022colorfilm}. Considering that the minuscule perturbations in restricted attacks are difficult to realize in physical scenarios (though one can optimize the perturbations and then print and paste images in the real world), it is more practical and convenient to launch the attack by changing some factors which are not related to the image semantics, \eg irrelevant object, environment, light. 

\subsection{Attacks against Face Anti-Spoofing}
Face anti-spoofing classification is the most crucial and leading stage in face recognition systems. We leave the detailed justification in Appendix~\ref{append:justification}. Apart from the general attacks mentioned above, we also provide an overview of the existing face anti-spoofing models. In a broad sense, face anti-spoofing models can be reckoned as a binary classification problem between ``bonafide'' and ``spoofing''. A plethora of DNN-based anti-spoofing models have sprung up recently, which are based on binary cross-entropy supervision~\cite{yang2019face,xu2021improving,wang2022patchnet} or pixel-wise supervision~\cite{atoum2017face,yu2020cdcn,yu2020face}. Pixel-wise supervision models can provide contextual information to help fine-grained feature learning, which is more effective. DepthNet~\cite{atoum2017face} first utilized pseudo-depth labels to supervise spoofed images with no facial depth information, unveiling the pixel-wise supervision models. Among the existing anti-spoofing models, CDCN++~\cite{yu2020cdcn}, as one of the most popular models~\cite{yu2022deep}, slightly modified DepthNet to Central Difference Convolution (CDC) to capture the central gradient of CNN sampling values. Meanwhile, a multiscale attention fusion module was designed to integrate CDC features at low, medium, and high levels. Following in the footsteps of CDCN++ were some by-products~\cite{yu2021dual,wu2021dual}. Despite some recent advanced anti-spoofing models that pursue domain generalization~\cite{liu2021adaptive,guo2022multi,wang2022domain,yue2023cyclically,zou2023adversarial,srivatsan2023flip}, they are not models on which we focus in this paper.

However, there is scanty research on adversarial attacks against face anti-spoofing models. Agarwal et al.~\cite{agarwal2019deceiving,agarwal2019deceiving2} first explored this area, followed by Zhang et al.~\cite{zhang2020adversarial}. But they only attacked handcrafted or shallow neural networks in a white-box setting. Bousnina et al.~\cite{bousnina2021unraveling} simulated adversarial attacks for transfer-based anti-spoofing models using differential evolution. Yang et al.~\cite{yang2023exposing} devised an augmentation method to make adversarial noise more semantic. As a whole, these works have not considered the black-box physical attack setting.

\subsection{Physical Attacks}
Some studies have begun to focus on physical attacks~\cite{athalye2018synthesizing,xu2020adversarial,sitawarin2018rogue}. Stop signs were attacked by stickers that look like graffiti~\cite{eykholt2018robust} or calculated adversarial perturbations produced by a projector~\cite{lovisotto2021slap}. The flickers were then simulated in the physically taken videos with Wifi-controlled RGB light bulbs~\cite{pony2021flickering}. In addition, colorization attacks, as an emergent branch of unrestricted attacks, maintain imperceptibility by changing the color of the image, without affecting the semantic information. Among them, a concurrent work, AdvCF~\cite{hu2022colorfilm}, utilized color film to create adversarial samples in the physical world, but the images taken lacked the depth information reflected by external light. Hoping to find imperceptible colors to deceive human eyes by altering different colors for various semantic classes, ColorFool~\cite{shamsabadi2020colorfool} and the lately proposed NCF~\cite{yuan2022natural} are unrealistic to mount in the physical world. Conversely, our face anti-spoofing task does not require high imperceptibility, \eg most real-world access control systems are unattended. 

Overall, adversarial perturbations in existing research are fully or partially relied on computers. That is to say, attackers are required to have extensive computer knowledge and programming capability when crafting adversarial perturbations. Nevertheless, we hope to expose a simple attack that even amateurs can implement.

\begin{figure*}[htb]
\begin{center}
  \includegraphics[width=.8\linewidth]{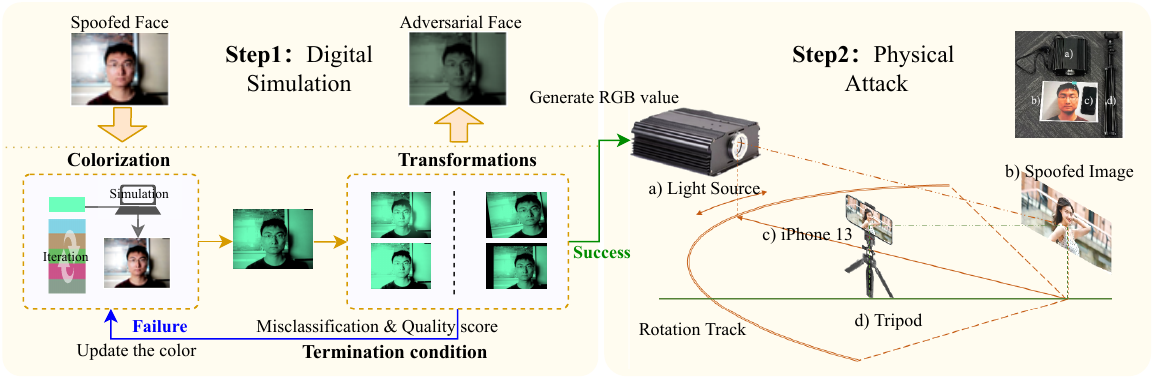}
  \vspace{-3mm}
\end{center}
  \caption{AdvColor framework.}
\label{fig:framework}
\vspace{-2mm}
\end{figure*}

\section{A Practically Accessible Attack}
To compare with the existing arsenal of attacks in our later study, we first propose an accessible but strong attack named AdvColor.

\subsection{Problem Definition}
The target face recognition pipeline is shown in Figure~\ref{fig:pipeline}. When an image is sent to the system, the face anti-spoofing classification result, face quality score and face matching ID are returned as the outputs. The attacker's aim is to evade the face recognition system via a printed portrait of a target person. Specifically, the system captures an image of the printed portrait and first sends it to the anti-spoofing model. A bonafide image with an above-threshold quality score will be sent to face matching for identity check. If the attack is rejected at any aforementioned stage, the attack is counted as a failure. In other words, a successful attack against face recognition systems (the adversarial face with green border in Figure~\ref{fig:pipeline}) should 1) alter a spoofing image to bonafide, 2) pass the face quality assessment, and 3) be matched to the identity of the target person to whom the portrait belongs.

However, in practice, we find that the face-matching module is extremely vulnerable once the attack can evade the face anti-spoofing model at a high face quality score. Therefore, we mainly focus on the definition of the face anti-spoofing model and the attack setting which are in detail elucidated as follows. Given a face anti-spoofing DNN model $f\left( {x} \right):x \to y$, it takes an image $x\in {\mathbb{R}^{H \times W \times C}}$ as input, where $H$, $W$, and $C$ are, respectively, the height, width, and channel number of $x$. The output of the model is binary, \ie $y \in \{0,1\}$, where 0 denotes spoofing and 1 denotes bonafide. The attack goal against the face anti-spoofing model is to find an adversarial version ${x_{adv}}$ for a spoofed image $x$ such that $f({x_{adv}}) = 1$. On the other hand, a quality assessment algorithm $Q\left( {x} \right):x \to s$ maps the image to a scalar score $s\in[0,1]$.

\subsection{Threat Model}
We focus on physical black-box settings due to their practicability. The attacker searches for the best adversarial RGB filters on a computer and instantiates them into adversarial illuminations in the physical world using ubiquitous and accessible devices. The illuminations create a persistently adversarial environment in which printed portrait photos can directly achieve the aforementioned attack purposes. Different from the commonly used soft-label setting in images, we adopt hard-label (a.k.a. label-only~\cite{ilyas2018black}) setting in our research, which is the most challenging black-box setting. 

\subsection{Proposed Approach}
Figure~\ref{fig:framework} outlines the framework of AdvColor. AdvColor first simulates adversarial RGB values by optimizing an adversarial color filter. Next, the RGB values are used as the specifications of a simple illumination device to turn human face photos into robust physical examples against black-box face recognition systems.

\noindent \textbf{Adversarial color filter.~}
In this paper, we consider adding perturbation by a color filter uniformly scaling the RGB channels of an image. Given the values of three channels of an image $R,G,B \in \left[ {0,255} \right] \subset {\mathbb{N}^{H \times W \times C}}$, we first grayscale the image so as to unify the color distribution.
{\small
\begin{equation}
gray = {0.3 \times R + 0.59 \times G + 0.11 \times B}, 
\end{equation}
}%
where $gray$ represents the gray value of the image.  
Given a color filter $\alpha_f  = {\left[ {{r_f},{g_f},{b_f}} \right]^T}$, where ${r_f}$, ${g_f}$, ${b_f}$ $\in \left[ {0,1} \right]$. Then the values of the $R$, $G$, $B$ channels of the filtered image can be expressed as
{\small
\begin{equation}\label{eq:color_filter}
\left\{ {\begin{array}{*{20}{l}}
{{R_f} = \lfloor \mathrm{clip} \left( {{r_f} \times gray,0,255} \right) \rfloor}, \\
{{G_f} = \lfloor \mathrm{clip} \left( {{g_f} \times gray,0,255} \right) \rfloor}, \\
{{B_f} = \lfloor \mathrm{clip} \left( {{b_f} \times gray,0,255} \right) \rfloor},
\end{array}} \right.
\end{equation}
}%
where ${R_f},{G_f},{B_f} \in \left[ {0,255} \right] \subset {\mathbb{N}^{H \times W \times C}}$ represents the values of three channels of the filtered image, $\mathrm{clip} $ limits the input within the given range, $\lfloor \cdot \rfloor$ denotes the floor operation. Although the pixel values of the image are all integers, they are normalized into floating point type before sending the image to the classifier. Therefore, we will remove the floor operation in Equation~\ref{eq:color_filter} in the following optimization stage unless otherwise stated.

\noindent \textbf{Color filter optimization.~}
Though white-box attacks can easily succeed if we take the color filter as optimization parameters and use gradient descent to launch the attack, it does not conform to reality. Generally, we cannot get the confidence score of the image from the public model, but we can only obtain the binary label. This requires us to consider the more difficult black-box setting.

To search for the color filter $\alpha_f $ which can fool the face anti-spoofing model, one substitute method is to use Particle Swarm Optimization (PSO)~\cite{eberhart1995pso} due to its efficiency in this low-dimensional optimization problem. Candidate color filters are regarded as the positions of particles. We uniformly initialize the particles' positions on $\left[ {0,1} \right]$, and limit the search space in this range to cope with the normalized pixel space. The fitness function can be expressed as
{\small
\begin{equation}\label{eq:AdvColor_one_image}
\mathop {\arg \min }\limits_{\alpha_f} \left [ {\mathds{1}\left( {f\left( {\Theta \left( {x,\alpha_f} \right)} \right) = 0} \right)} \right],
\end{equation}
}%
where $\mathds{1}(\cdot)$ refers to the indicator function, and $\Theta(\cdot)$ represents the colorization transformation on an image $x$ given $\alpha_f$, as described in Equation~\ref{eq:color_filter}. We provide more explanation for why we choose PSO to update colors in Section~\ref{subsec:pso}.

\noindent \textbf{Adaptation to physical attacks.~}
The colorized image generated by Equation~\ref{eq:AdvColor_one_image} has an obvious disadvantage, \ie the image looks unnatural compared to photos taken physically under RGB light sources, especially for bright colors. Moreover, the generated image may not be robust enough when applied to physical-world attacks. In order to simulate the environment of physical scenes, we introduce various transformations to improve the robustness of the colorized images. Photos taken in the physical world are affected by many external factors, \eg shooting angle, illumination distance, and light intensity. Concretely, we consider illumination, brightness, gamma correlation, and other transformations during the optimization.

\begin{itemize}
\item Illumination.
Halation occurs when taking photos in real-world scenarios.
To simulate this halation effect, we randomly select an illumination center $\left( {{x_c},{y_c}} \right)$ in the image and update the RGB pixels within the illumination range as follows.
{\small
\begin{equation}
\left\{ {\begin{array}{*{20}{c}}
{{R_{ij}} \leftarrow \mathrm{clip} \left( {R_{ij}} + \kappa \max \left( {1 - \frac{{{d_{ij}}}}{{{R_c}}},0} \right), 0, 255 \right)}, \\
{{G_{ij}} \leftarrow \mathrm{clip} \left( {G_{ij}} + \kappa \max \left( {1 - \frac{{{d_{ij}}}}{{{R_c}}},0} \right), 0, 255 \right)}, \\
{{B_{ij}} \leftarrow \mathrm{clip} \left( {B_{ij}} + \kappa \max \left( {1 - \frac{{{d_{ij}}}}{{{R_c}}},0} \right), 0, 255 \right)},
\end{array}} \right.
\end{equation}
}%
where $R_{ij}$, $G_{ij}$ and $B_{ij}$ represents the $R$, $G$ and $B$ pixel values in the position $\left( i,j \right)$, $\kappa $ is the luminance coefficient, ${d_{ij}}$ stands for the Euclidean distance between $\left( i,j \right)$ and the illumination center $\left( {{x_c},{y_c}} \right)$, $R_c$ is the illumination radius.

\item Brightness.
The brightness of a physical photo will be affected by daylight and camera settings, \eg a photo taken near the window is brighter; while the one taken at night is darker. Therefore, we apply different brightness adjustments to the colorized image, which can help us obtain a wider range of tones. We adjust the image brightness by $x \leftarrow \mathrm{clip} \left( {\lambda x,0,255} \right)$, where $\lambda $ is the brightness coefficient. 

\item Gamma correlation.
There is a discrepancy between a physical scene and a photo of the scene when interpreted by human eyes, \ie the brightness in nature does not change uniformly, while the brightness seen and felt by human eyes changes uniformly. Therefore, based on this visual deviation, it is usually necessary to perform the Gamma correction on the image after the brightness adjustment. Specifically, the image is slightly modified by $x \leftarrow \mathrm{clip} \left( \sqrt[r]{x}, 0, 255 \right)$, 
where $\gamma$ is the Gamma correlation coefficient.

\item Other transformations.
We additionally consider random translation, rotation, cropping, and Gaussian blurring to simulate the actual shooting scenes.
\end{itemize}

In order to simulate physical attacks, the Expectation Over Transformation (EOT)~\cite{eykholt2018robust} is used here to minimize the object function over different transformations. The input image is first colorized, and then transformed by a composition of various transformations mentioned above during the adversarial attack. Since halation does not always exist, we introduce a probability to decide whether to perform the illumination transformation. Furthermore, we hope that the attack can be completed without reducing the face quality score, a metric to evaluate the utility, fidelity, and character of the face image~\cite{schlett2022face}. In this paper, we use one of the most advanced models, namely LightQNet~\cite{chen2021lightqnet}, to evaluate face quality. To maintain the face quality score, the fitness function is set as follows:
{\small
\begin{equation}\label{eq:AdvColor_both}
\mathop {\arg \min }\limits_{{\alpha _f}} {\mathbb{E}_{t \sim T}}\left\{ {\mathds{1}\left( {f\left( {{x_c}} \right) = 0} \right) + \lambda \left[ {\left( {1 - Q\left( {{x_c}} \right)} \right)\mathds{1}\left( {f\left( {{x_c}} \right) = 1} \right)} \right]} \right\},
\end{equation}
}%
where $T$ is the set of all transformations, ${x_c} = t\left( {\Theta\left( {x,{\alpha _f}} \right)} \right)$ is the image after colorization and a random transformation, $Q\left ( \cdot \right)\in[0,1]$ is the quality score of a given image, and $\lambda $ is a constant balancing the two objectives. Minimizing the above objective maximizes the probability of misclassifying the image from spoofing (label 0) to bonafide (label 1) while keeping the quality score as high as possible. Solving Equation~\ref{eq:AdvColor_both} gives the optimal adversarial color filter $\alpha _f$.

\subsection{Overview of AdvColor}
The whole process of AdvColor is shown in Algorithm~\ref{alg:AdvColor_attack}. First, an input image $x$ is colorized and transformed. The fitness function is then calculated, and the color that can misclassify the input image under various transformations and maintain the quality score is optimized by using PSO. Finally, the adversarial image $x_{adv}$ is obtained. Note that the quality score may be reduced while increasing the misclassification rate. Since performing the line search in Equation~\ref{eq:AdvColor_both} could be time-consuming, we use an elastic search algorithm in the optimization. Specifically, the optimization stops when ${\mathbb{E}_{t \sim T}} {\mathds{1}\left( {f\left( {{x_c}} \right) = 0} \right)} \leq 0.1$ and ${\mathbb{E}_{t \sim T}} \left[ \left( 1- {Q\left( {{x_c}} \right)}\right) \mathds{1}\left( {f\left( {{x_c}} \right) = 1}\right)\right]\geq{0.5}$ are met simultaneously. On the other hand, since PSO may fall into local optimums when the optimal fitness function does not change in ten consecutive iterations, we also cease the optimization.

\noindent
\begin{algorithm}[t]
\footnotesize
\caption{AdvColor.}\label{alg:AdvColor_attack}
\KwIn{Black-box classifier $f$, input image ${x}$, target class ${y_t}$, sample number $N_s$, color filter $\alpha_f$, particle number $N_p$, sequential composition $T$.
}
\KwOut{Adversarial image ${x_{adv}}$.}
\While{not meeting the termination condition}{
    $err,err_q \leftarrow 0, 0$;\\
    \For {$k \gets 1$ to $N_s$}{
        $t \leftarrow$ a random sample from $T$;
        ${x_{adv}} \leftarrow t\left( {\Theta\left( {x,{\alpha _f}} \right)} \right)$;\\
        \If{$f\left( {{x_{adv}}} \right) =  = {y_t}$}
        {
            $err \leftarrow err + 1$;\\
            $err_q \leftarrow err_q + 1 - Q\left( {{x_{adv}}} \right)$;
        }
    }
    $fitness \leftarrow 1 - \frac{{err}}{{{N_s}}} + \lambda \frac{{er{r_q}}}{{err}}$;\\
    $\alpha_f \leftarrow PSO\left ( {fitness,{N_p}} \right )$;
}
return one of the misclassified $x_{adv}$.
\end{algorithm}

\vspace{-6mm}

\section{Machine Experiments}\label{sec:experiments}
In this section, we benchmark AdvColor with several state-of-the-art and sophisticated (\eg utilizing gradient estimation, heuristic search, \etc) attacks to show the capability of AdvColor in fooling machines. We conduct the benchmarking in both digital and physical settings, since adversarial attacks are usually delivered directly in digital pipelines or through realized physical adversarial examples. Moreover, we investigate the capability of AdvColor in bypassing adversarial defenses.

\subsection{Experimental Details}
\noindent \textbf{Dataset and models.}
We use Replay-Attack~\cite{chingovska2012REPLAY-ATTACK} and CDCN++~\cite{yu2020cdcn} as our dataset and victim face anti-spoofing model due to their representativeness and popularity. As one of the most popular face anti-spoofing datasets, Replay-Attack comprised of 1,300 videos of 50 different subjects under print, mobile replay and high-definition replay. We split the dataset into a training set and a test set with a proportion of 7:3. As one of the most widely used face anti-spoofing models, CDCN++ is formed by three feature extraction blocks and a multi-head attention module. CDCN++ with depth supervision is recognized as the most excellent architecture due to its compactness and classification performance~\cite{yu2022deep}. The pre-trained model achieves 98\% AUC (Area Under the Curve) on test set. Then, we use a pre-trained LightQNet~\cite{chen2021lightqnet} to evaluate the face quality score\footnote{https://github.com/KaenChan/lightqnet}. Finally, a pre-trained face identity matching model is used for face matching\footnote{https://github.com/ageitgey/face\_recognition}.

\noindent \textbf{Evaluation metrics.}
In the black-box attack, we use the following metrics to evaluate the attack performance.
1) \textbf{Fooling Rate (FR)}: the ratio of spoofed images misclassified as bonafide within the given query limit. In order to evaluate the efficiency of different attack methods, we set a query limit to avoid redundant queries, particularly for competitors with low efficiency. FR resembles the Attack Presentation Classification Error Rate (APCER), which is commonly used in face anti-spoofing classification tasks.
2) \textbf{Average Query (AQ)}: the average query number for a successful attack.
3) \textbf{Average Quality Score (AQS)}: the average face quality score~\cite{chen2021lightqnet}.
4) \textbf{Overall Attack Success Rate (OASR)}: the ratio of images that are matched to correct identities to those that can successfully evade the anti-spoofing model and pass the face quality assessment. In other words, OASR gives the overall attack success rate against the face recognition system.
5) \textbf{Adversariality}: the ratio of spoofed images that are misclassified as bonafide after transformations sampled from $T$. The adversariality of a clean image $x$ is defined as ${\mathbb{E}_{t \sim T}}\left[ {\mathds{1}\left( {f\left( {{t}\left( x \right)} \right) = 1} \right)} \right]$. Similarly, an adversarial image has an adversariality of ${\mathbb{E}_{t \sim T}}\left[ {\mathds{1}\left( {f\left( t\left( {\Theta\left( {x,{\alpha _f}} \right)} \right) \right) = 1} \right)} \right]$. We use \textbf{Adv-m} and \textbf{Adv-s} to record the mean and standard variance of adversariality over all samples. \textbf{Adv-m} describes the overall attack effectiveness and \textbf{Adv-s} tells how stable the attack is.

\noindent \textbf{Competitors.}
Since there is no existing work studying black-box physical attacks in face anti-spoofing tasks, we adopt four prevailing hard-label black-box attacks in image classification tasks, namely Ilyas~\cite{ilyas2018black}, SignOpt~\cite{cheng2020signopt}, RayS~\cite{chen2020rays} and TA~\cite{wang2022triangle}, and extend them to face anti-spoofing tasks. For a fair comparison, we conduct experiments on four competitors and AdvColor without applying transformations. Furthermore, we evaluate AdvColor under various transformations to infer its performance in physical scenes. The comparison with existing color attacks will be discussed in Section~\ref{discuss:other_color_attacks}.

\noindent \textbf{Fairness.}
Obviously, it is not fair to compare AdvColor with competitors with restricted attacks. As a remedy, we slightly modify competitors to unrestricted attacks by enlarging the $\ell_p$ bound of competitors to the same $\ell_p$ magnitudes of AdvColor. We provide both $\ell_2$ and $\ell_\infty$ versions of competitors due to their scalability and use their default parameters. In this way, we unify the adversary budget of AdvColor and that of the others at the same level. For AdvColor, it is degraded to another variant, AdvColor-AS, which does not consider various transformations and only attacks the anti-spoofing model (namely $ \lambda=0 $ in Equation~\ref{eq:AdvColor_both}) to ensure fairness in comparison with other competitors. We set the above comparison principles also based on observations that the competitors all fail when the transformations are applied. The adversarial samples found by the competitors are almost in the vicinity of the decision boundary, especially for decision-based attacks, \ie SignOpt~\cite{cheng2020signopt} and RayS~\cite{chen2020rays}. However, these attacks cannot converge when transformations such as illumination and rotation are employed. This phenomenon occurs commonly when EOT is applied to competitors, leading to low adversariality. Therefore, we can only degrade AdvColor to fit the fair comparison with the competitors.

\begin{table}[tb]  
\centering
\caption{Results of AdvColor using different values of $\lambda $.}
\label{tab:grid_search}
\resizebox{.8\linewidth}{!}{
\begin{tabular}{c|ccccccccc}
\toprule
\textbf{$\lambda $} & \textbf{FR}$\uparrow$ & \textbf{AQ}$\downarrow$ & \textbf{Adv-m}$\uparrow$ & \textbf{AQS}$\uparrow$ & \textbf{OASR}$\uparrow$ \\
\midrule
0 & 100\% & 21,547 & 91.3\% & 0.3326 & 29\% \\
0.2 & 100\% & 20,335 & 88.5\% & 0.4791 & 78\% \\
0.4 & 100\% & 19,196 & 83.0\% & 0.5105 & 82\% \\
0.6 & 100\% & 19,227 & 85.2\% & 0.5644 & 94\% \\
0.8 & 100\% & 19,465 & 82.5\% & 0.5752 & 94\% \\
1.0 & 100\% & 19,193 & 81.0\% & 0.5802 & 95\% \\
\bottomrule
\end{tabular}
}
\end{table}

\begin{table}[t]  
\centering
\caption{Reference values of transformation parameters.}
\label{tab:transformation_parameters}
\vspace{1mm}
\resizebox{0.95\linewidth}{!}{
\begin{tabular}{cc|cc}
\toprule
\textbf{Parameter} & \textbf{Range} & \textbf{Parameter} & \textbf{Range}\\
\midrule
Illumination coefficient & $\left[ {0,300} \right]$ & Translation & $\left[ {-10,10} \right]$ \\
Illumination center & $\left[ {25,100} \right]$ & Rotation & $\left[ {-60^\circ ,60^\circ } \right]$\\
Illumination radius & $\left[ {25,50} \right]$ & Crop & $\left[ {-20,20} \right]$ \\
Brightness coefficient & $\left[ {0.2,1.8} \right]$ & Gaussian kernel size & $\left\{ {3,5,7} \right\}$ \\
Gamma correlation & $\left[ {1,3} \right]$  \\
\bottomrule
\end{tabular}
}
\end{table}

\noindent \textbf{Parameter settings.}
To balance the attack performance and retain the face quality, we perform grid search on 50 randomly selected images to identify the proper value of $\lambda $. We select $\lambda $ from $\{ 0, 0.2, 0.4, 0.6, 0.8, 1.0 \}$. Table~\ref{tab:grid_search} lists the attack performance. It can be observed that larger $\lambda$ may result in low adversariality and lower $\lambda $ reduces the quality score. Considering the five metrics comprehensively, we finally set $\lambda=0.6$. Note that the case where $\lambda = 0$ is different from AdvColor-AS. Here we consider various transformations and attacking the face recognition system, while AdvColor-AS only attacks the anti-spoofing model (\ie not considering maintaining the face quality) and does not consider transformations in order to launch fair comparison with competitors.

We set the transformation sampling number $N_s = 40$ and the particle number $N_p = 30$. The maximum iteration number in PSO is set to 100. Table~\ref{tab:transformation_parameters} provides the ranges of reference values for parameters used in various transformations. All parameters in the experiments are uniformly sampled from the given ranges. We set the query limit as 10,000, since we find that most of the competitors can complete attacks within this upper limit in our experiments.

\subsection{Experimental Results on Simulation Attacks}
\noindent \textbf{Attack performance.}
We randomly select 200 images whose ground truth labels are spoofing from the Replay-Attack dataset and craft their adversarial versions against the black-box CDCN++ model using AdvColor and all the competitors. We first compare the attack performance against anti-spoofing models when no transformation is considered. The results reported in Table~\ref{tab:quantitative_analysis} show that AdvColor-AS performs better than other competitors in all metrics. In particular, AdvColor-AS has an average query of 34, which is much lower than that of competitors. Although we equalize the $\ell_p$ bound between AdvColor and other competitors, the fooling rate of competitors does not increase significantly with the number of queries. This is because searching for perturbations under large $\ell_p$ bounds takes a much greater effort to converge. When imperceptibility is not a primary concern, the adversary can save a tremendous number of queries, as demonstrated by AdvColor-AS. The competitor method with the closest performance to AdvColor-AS is RayS. Since RayS introduces the fast hierarchical search and adds perturbations horizontally, it can achieve relatively good Adv-m when not considering various transformations. However, the attack performance is comprehensively determined by various metrics; the OASR is still low enough that the attack ability towards the overall face recognition pipeline is limited. Also, RayS's fooling rate under the same query number does not exceed that of AdvColor-AS. Other competitors, especially SignOpt, have poor adversariality scores. Since the adversarial examples found reside in the vicinity of the decision boundaries, a random transformation may easily invalidate the adversarial examples. When considering the results of the face quality assessment, AdvColor-w/o performs better than AdvColor-AS in terms of AQS and OASR. When additionally considering transformations, AdvColor consumes an average query of 24,315 while maintaining $100\%$ FR and 94\% OASR, indicating that the face recognition system is wholly paralyzed. In contrast, all the competitors fail to attack the anti-spoofing model when the transformations are considered, let alone subsequent stages (FR=0).

\begin{table}[t]  
\centering
\caption{Attack simulation performance under the hard-label black-box setting (w/o: Without EoT; w/: With EoT).}
\vspace{-1mm}
\label{tab:quantitative_analysis}
\resizebox{0.9\linewidth}{!}{
\begin{tabular}{ccccccccrrr}
\toprule
\textbf{Method} & \textbf{Tran} & \textbf{FR}$\uparrow$ & \textbf{AQ}$\downarrow$ & \textbf{Adv-m}$\uparrow$ & \textbf{AQS}$\uparrow$  & \textbf{OASR}$\uparrow$ \\
\midrule
Ilyas (unres,$L_{\inf}$)~\cite{ilyas2018black} & w/o & 100\% & 1,589 & 28\% & 0.2352 & 1\%\\
Ilyas (unres,$L_2$)~\cite{ilyas2018black} & w/o & 79\% & $>$2,746 & 26\% & 0.2998 & 1\%\\
RayS (unres,$L_{\inf}$)~\cite{chen2020rays} & w/o & 100\% & 189 & 42\% & 0.2266 & 29\% \\
RayS (unres,$L_2$)~\cite{chen2020rays} & w/o & 100\% & 320 & 38\% & 0.3643 & 47\% \\
SignOpt (unres,$L_2$)~\cite{cheng2020signopt} & w/o & 100\% & 1,536 & 3\% & 0.1853 & 12\% \\
SignOpt (unres,$L_{\inf}$)~\cite{cheng2020signopt} & w/o & 58\% & $>$5,285 & 2\% & 0.1725 & 6\% \\
TA (unres,$L_2$)~\cite{wang2022triangle} & w/o & 100\% & 329 & 47\% & 0.1796 & 16\% \\
TA (unres,$L_{\inf}$)~\cite{wang2022triangle} & w/o & 26\% & $>$7,929 & 17\% & 0.1423 & 6\% \\
AdvColor-AS & w/o & 100\% & \textbf{34} & \textbf{49}\% & \textbf{0.3814} & \textbf{72}\% \\
\midrule
AdvColor-w/o & w/o & 100\% & \textbf{76} & 53\% & 0.5652 & \textbf{97\%} \\
AdvColor & w/ & 100\% & 24,315 & \textbf{87\%} & \textbf{0.5946} & 94\%\\
\bottomrule
\end{tabular}
}
\end{table}

\begin{table}[t]
\centering
\caption{Quantitative statistics.}
\vspace{-1mm}
\label{tab:adversariality_and_quality_and_matching_performance}
\resizebox{0.8\linewidth}{!}{
\begin{tabular}{ccccccccc}
\toprule

& \multicolumn{2}{c}{\textbf{Adversariality}} & \multicolumn{2}{c}{\textbf{Face quality}} & \textbf{Face recognition} \\
\cmidrule(r){2-3}\cmidrule(r){4-5}\cmidrule(r){6-6}
& \textbf{Adv-m}$\uparrow$ & \textbf{Adv-s}$\downarrow$ & \textbf{VIF}$\uparrow$ & \textbf{AQS}$\uparrow$ & \textbf{OASR}$\uparrow$ \\
\midrule
Clean & 0.32\% & 0.83\% & \multirow{2}{*}[-0.5ex]{0.8338} & 0.6062 & -- \\
Adversarial & 86.76\% & 3.80\% & & 0.5946 & 94.38\% \\
\bottomrule
\end{tabular}
}
\end{table}

\begin{figure}[t]
  \centering
  \includegraphics[width=1.0\linewidth]{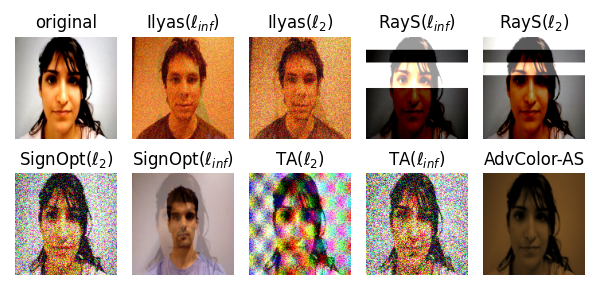}
  \vspace{-6mm}
   \caption{Visualizations of an original image and different adversarial images. 
   }
   \label{fig:vis_compare}
   \vspace{-3mm}
\end{figure}

From another crucial perspective, we note that competitors cannot preserve the face quality and semantics of the faces, as shown in Figure~\ref{fig:vis_compare}. Since the $\ell_p$ bound is relaxed, palpable perturbations appear in the face region, which consequently lead to the futility of the attack against the face recognition system, since the examples crafted by the competitors cannot pass the face quality assessment and face ID matching. On the contrary, our method can launch successful attacks against the whole system by inducing non-semantic changes to the faces. As shown in Table~\ref{tab:adversariality_and_quality_and_matching_performance}, the adversarial images after attacks have a mean adversariality of 86.76\%, which indicates that AdvColor can achieve a very high misclassification rate under different transformations and additionally reflects the effectiveness of AdvColor in generating robust adversarial samples. We further evaluate the face quality between clean images and adversarial images by \emph{Visual Information Fidelity} (VIF)~\cite{sheikh2006image} and AQS. VIF measures the semantic similarity between images with non-semantic but large perturbations. The VIF between images before and after attacks is 0.8338, which is high enough to show that the semantic information of the AdvColor examples is well preserved. Since the fitness function in Equation~\ref{eq:AdvColor_both} considers the trade-off between misclassification and retaining face quality, adversarial images still maintain an AQS of 0.5946 (compared to 0.6062 of spoofed images before attacks). AdvColor only leads to a marginal decrease in AQS. After face matching, we find that over $94\%$ of adversarial images can be correctly matched with the correct face identity, which indicates that AdvColor is strong enough to paralyze the entire face recognition system. Therefore, AdvColor can achieve high adversariality, maintain face quality, and culminate in failure of the face recognition pipeline. 

In summary, AdvColor-AS only attacks the anti-spoofing model to ensure fair comparison. AdvColor-w/o additionally considers the face quality score in optimization. Since colorization will not markedly affect the face quality score, the AQ is just doubled, but the AQS and OASR improves a lot. Therefore, in digital attacks, AdvColor-w/o is more recommended. AdvColor additionally adds various transformations. The AQ rises since transformations increase the attack difficulty, but the Adv-m and AQS improves a lot. Despite higher AQ and slightly fewer OASR, the color filters obtained are better since they can directly adapt to physical attacks.

\noindent \textbf{Visualization.}
Although the access control system may sit in unattended situations in which the perceptibility of the adversarial attack is not a major concern, the quality and semantics of the face images need to be guaranteed basically, since these greatly affect the passing rate of face quality assessment and the correctness of face matching. Figure~\ref{fig:vis_compare} visualizes a captured original spoofing image (replayed on a high definition screen) and adversarial images (misclassified as bonafide) generated by different attacks. Unfortunately, we find that, when using the $\ell_p$ bound equal to that of AdvColor, adversarial images of competitors contain large amounts of irregular and anomalous perturbations. As Ilyas~\cite{ilyas2018black} and SignOpt~\cite{cheng2020signopt} initialize the search with a bonafide image chosen from the bonafide class, the output adversarial images are totally different from the original image due to large $\ell_p$ bound. This largely reduces the matching correctness in face matching. RayS~\cite{chen2020rays} initializes the search direction as 1 and adds perturbations horizontally. When the $\ell_p$ bound is large, the attack terminates early and the adversarial example exhibits abnormal light in the face area, which on a large scale affects the face quality assessment and face matching results since the face region becomes hard to detect. TA searches for the candidate triangle and adjusts the angle iteratively to find adversarial images. However, abnormal patterns appear in the output adversarial image, which reduces face quality by a large margin. This unexpected phenomenon also exists in their original paper when the RMSE is large~\cite{wang2022triangle}. Conversely, the non-semantic perturbations added in our method are smooth enough to maintain the face information, so as to pass the face quality assessment and make face matching output the same ID as that of the original image. 

\begin{figure}[t]
\begin{center}
  \includegraphics[width=0.75\linewidth]{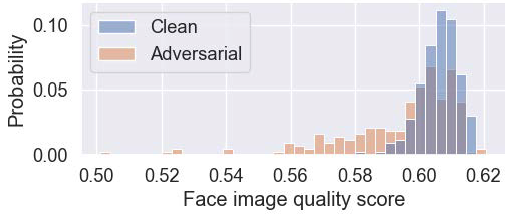}
\end{center}
\vspace{-2mm}
  \caption{Distributions of quality scores before/after attacks.}
\label{fig:quality_score_distribution}
\vspace{-3mm}
\end{figure}

\noindent \textbf{Quality score distribution.}
Figure~\ref{fig:quality_score_distribution} shows the distribution of the quality score of clean images and adversarial images for better visualization. For the adversarial images, most of the scores are just slightly reduced. A surprising discovery is that the face quality of few adversarial images even outnumbers that of the original clean images. One reasonable explanation for this phenomenon could be the increase in illumination and brightness introduced in random transformations.

\begin{figure}[t]
\begin{center}
  \includegraphics[width=.75\linewidth]{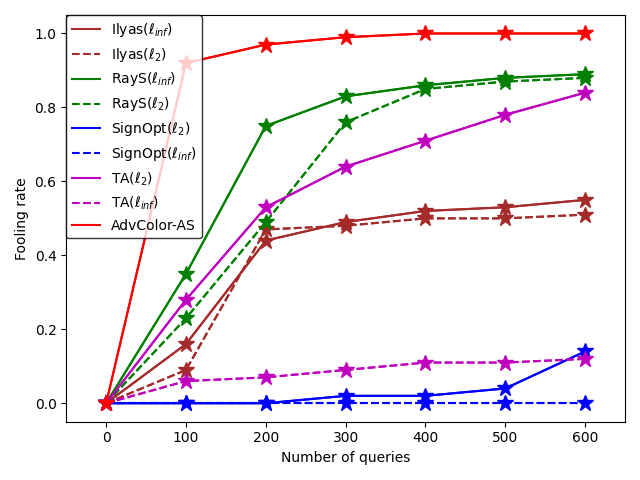}
\end{center}
\vspace{-3mm}
  \caption{Fooling rate w.r.t. number of queries under different attacks.}
\label{fig:fr&query}
\vspace{-3mm}
\end{figure}

\noindent \textbf{Other analyses.}
Figure~\ref{fig:fr&query} shows the comparison results of fooling rate \textit{w.r.t.} number of queries among all competitors and AdvColor-AS. The star stands for the fooling rate against the face anti-spoofing model when the number of upper queries is limited. Since Ilyas~\cite{ilyas2018black} and RayS~\cite{chen2020rays} are tailored for $\ell_\infty$ perturbations, while SignOpt~\cite{cheng2020signopt} and TA~\cite{wang2022triangle} perform better under $\ell_2$ perturbations, we keep their original settings and use their default parameters except for enlarging the $\ell_p$ bound. In addition, we also provide another $\ell_p$ version for comparison due to their scalability. The original versions are denoted as solid lines, whereas the extended versions are denoted as dashed lines in the figure. In general, the original versions perform better than their extended versions when the number of queries increases, which is in line with anticipation due to their specificity. We find that SignOpt and Ilyas need relatively more queries to achieve high fooling rates, indicating that they are not query-efficient under the restriction of the large $\ell_p$ bounds. TA($\ell_2$) performs significantly better than TA($\ell_\infty$). One possible reason could be that the triangle optimization theory is mainly based on the Euclidean distance between two samples. Owning to the fast hierarchical search, both RayS($\ell_\infty$) and RayS($\ell_2$) secure a good fooling performance. Their query efficiency, however, still falls behind that of AdvColor-AS, since AdvColor-AS achieves a more than 90\% fooling rate when the query limit is 100. Besides, the weakness of competitors can also be exposed by the low passing rates of the face quality score assessment and the low overall attack success rates against the face recognition system.

\begin{figure}[t]
\begin{center}
  \includegraphics[width=.75\linewidth]{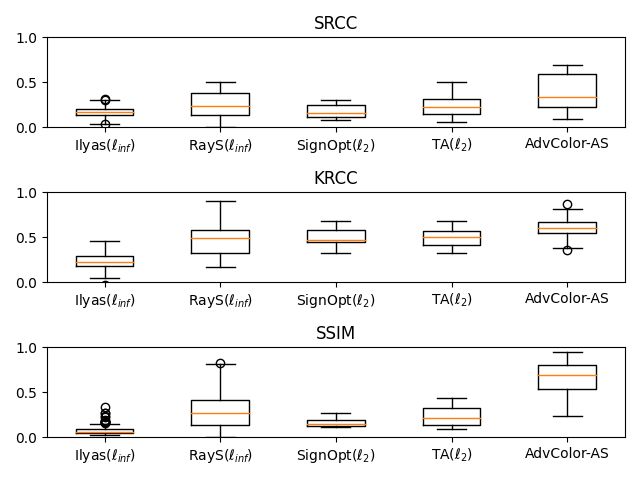}
\end{center}
\vspace{-3mm}
  \caption{Boxplots for image quality comparison (the higher the better). }
\label{fig:image_quality_comparison}
\end{figure}

\begin{figure}[t]
\begin{center}
  \includegraphics[width=.75\linewidth]{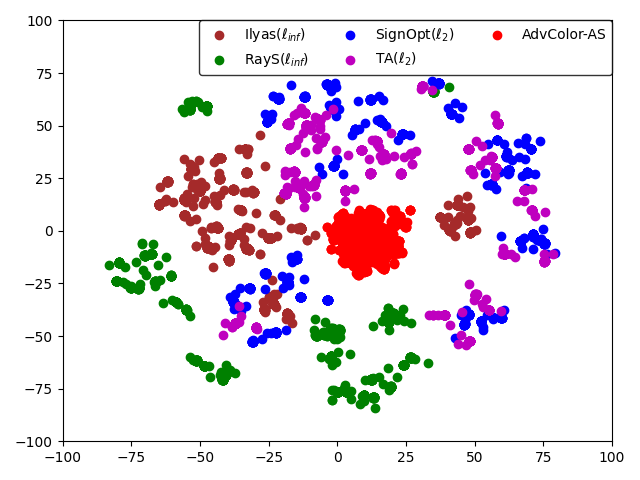}
\end{center}
\vspace{-3mm}
  \caption{t-SNE visualization of adversarial samples generated by different attacks.}
\label{fig:tsne}
\end{figure}

From the experimental perspective, we deem that when the $\ell_p$ bound increases for competitors, it is hard for them to find adversarial examples located in the same region as that of AdvColor-AS. Although the $\ell_p$ bound is expanded, the competitors still follow their original methodology in the search. For example, Ilyas~\cite{ilyas2018black} continuously reduces the $\ell_p$ distance between the adversarial image in the target class and the original image while maintaining the class of the adversarial image. SignOpt~\cite{cheng2020signopt} searches for the direction perpendicular to the decision boundary and looks for adversarial examples around the boundary. Given a large $\ell_p$ bound, the examples found by the competitors usually contain non-smooth perturbations, and the semantic information of the original images is hardly retained. However, AdvColor-AS modifies only the examples through color filters. Its perturbation preserves the semantic information of the image. For quantitative analyses, we calculate the image quality correlation between the adversarial examples and the original examples. As shown in Figure~\ref{fig:image_quality_comparison}, AdvColor-AS outperforms all the competitors in terms of three image quality metrics, SRCC~\cite{sheskin2007spearman}, KRCC~\cite{abdi2007kendall} and SSIM~\cite{wang2004ssim}. The results indicate that AdvColor-AS can preserve image quality well when considering unrestricted perturbations. Also, the methodological uniqueness leads to that the adversarial examples generated by AdvColor-AS obey the same data distribution in the hyperspace and this distribution is totally different from that of adversarial examples generated by the competitors. Figure~\ref{fig:tsne} shows the t-SNE visualization~\cite{van2008visualizing} of adversarial samples generated by competitors and AdvColor-AS. The distribution of examples generated by the competitors is highly distinct from that of AdvColor-AS. The above experimental results show that the examples found by AdvColor-AS are located in a uniform distribution where good image quality is maintained. Meanwhile, they ensure a high attack success rate and attack efficiency against the overall face recognition system. Although the AdvColor-AS samples are a subset of the points on an $\ell_{\infty}$ ball, we believe that competitors find it difficult to reach the corresponding regions. We point out that this gap is caused by methodological differences.

\subsection{Experimental Results on Physical Attacks}
\noindent \textbf{Attack performance.}
We employ a 45-Watts LED optical fiber machine to produce light source of any given RGB specifications, for the \emph{evaluation purpose}. In practice, any devices with tunable RGB values (\eg smart light bulbs, downlights) can be used to launch the attack. We capture photos of the physical images with an iPhone 13 camera in its default settings. The iPhone is fixed on a tripod. The right part of Figure~\ref{fig:framework} illustrates the setting of the scene when we take photos. The devices include a) an RGB light source, b) a print of an spoofed image, c) an iPhone 13 and d) a tripod. The light source and the tripod cost less than \$38 in total. We print the original spoofed images and take the photo of the print in an open space. The distance between the camera and the print ensures that the face area occupies the entire phone screen. The FR of the original photos is only 7\%, which indicates that the classifier performs well against normal prints. Then, we turn on the light source with different colors and take photos of the colorized prints with varying illuminating angles, shooting angles, and illumination distances. Finally, these photos are sent to the computer to test the attack performance on the victim model.

\begin{figure}[t]
\begin{center}
    \includegraphics[width=.98\linewidth]{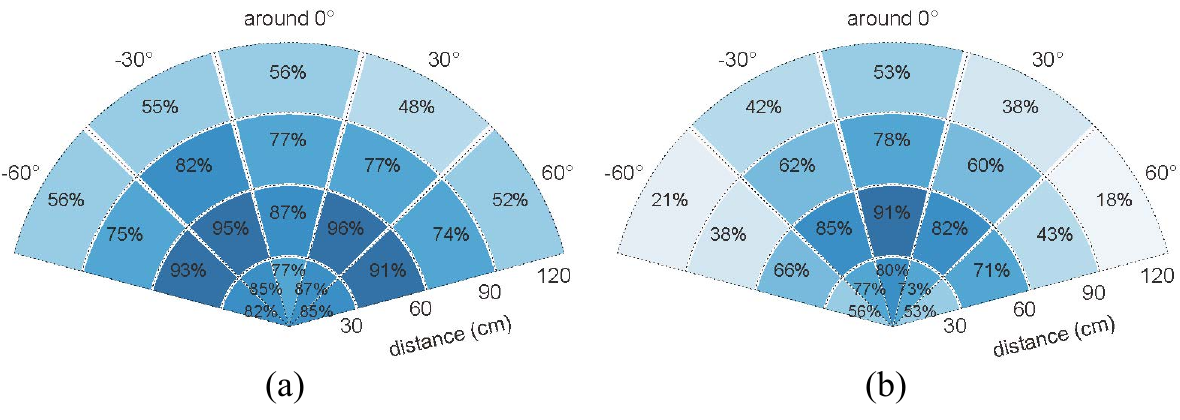}
\end{center}
\vspace{-3mm}
  \caption{Fooling rates under different angles and distances in physical scenes. (a) $\pm 60^\circ$, $\pm 30^\circ$, and around $0^\circ$ ($0^\circ$ excluded) refer to the angles between the light source direction and the face print. The shooting direction is perpendicular to the face print. (b) $\pm 60^\circ$, $\pm 30^\circ$, and around $0^\circ$ ($0^\circ$ excluded) refer to the angles between the shooting direction and the face print. The light source direction is perpendicular to the face print.}
\label{fig:physical_performance}
\end{figure}

Figure~\ref{fig:physical_performance} reports the performance of the physical attack at different angles and distances. Notice that a shooting angle of $0^\circ$ cannot be achieved due to the fact that the light source and the camera would block each other when taking the photo of the face prints. We therefore omit $0^\circ$ from the evaluation. The ``around $0^\circ$'' area represents a random selection of the angle from $\left [-30^\circ, 30^\circ \right]\setminus0^\circ$. The results show that the attack performs the best when the distance is 60 cm since a closer distance brings dazzling reflection of light, and a farther distance leads to darker photos. In terms of the light source angle, the success rates at $30^\circ$ and $60^\circ$ are higher than those around $0^\circ$, which is attributed to the intense reflection of light around $0^\circ$. The settings of $30^\circ$ and $60^\circ$ also align well with the actual attack situation. With regard to the shooting angle, we find that smaller angles result in higher fooling rate, owing to less face deformation. Furthermore, in the best shooting scene, the adversarial photos achieve an FR of 96\%, an AQS of 0.5825 and an OASR of $88\%$, which are close to those of the simulated attacks. This indicates that the face recognition system can be paralyzed with ease by AdvColor. 

\noindent \textbf{Mobile device.}
We use the iPhone as the shooting device to test the physical attack since it is one of the most widely used mobile phones. Since different mobile phones may result in performance discrepancy under the same attack. Therefore, we conduct additional experiments on another Android phone. We do not find a large difference between the attack results of the Android phone and the iPhone, since the fooling rate is also around 75\% to 95\% when the light source angles and shooting angles are within $\pm 60^\circ$ and the shooting distance is shorter than 60 cm. We attribute the similarity of attack performance on different devices to the power of color illumination, which can produce forceful implications for the face models.

\noindent \textbf{Attacks on business models.}
It would be difficult to access business models since most of them lead to great expense when conducting large-scale attacks or even hinder access by limiting the number of user visits. Despite the difficulty, we try our best to conduct small-scale sample attacks on two face anti-spoofing business models. Since it is unrealistic to optimize the color filter for different samples on the models due to the access limitation, we take 500 valid photos under different colors of illumination in the physical scenarios. The colors are selected randomly, and the light source angle, shooting angle and shooting distance are also randomly selected within the range in Figure~\ref{fig:physical_performance}. We find that the fooling rates are 10.6\% and 3.2\% for Company A and B respectively. Attacks on commercial models are more difficult because, in addition to the anti-spoofing model itself, there are often many external limitations that make the captured images invalid. For example, it is necessary to include the surrounding parts of the face in the photo, rather than only capturing the face part, which will capture the edge details of the printed paper, therefore reducing the attack performance. Another example is that Company B's model is very sensitive to the lighting at the shooting site, which often leads to the failure of face images captured in colored light before accessing the model. Since our proposed attack only focuses on exposing the vulnerability of the anti-spoofing model itself, we argue that how to bypass those strict limitations prior to the anti-spoofing detection in the business models is out of the scope of this paper. We will make it a part of our future work.

\subsection{Experimental Results on Universal Attacks}
In the simulated attack, we find that the optimized color filters of a large number of the adversarial examples converge to several similar color patterns (\eg purple in our case). We conjecture that the model might be sensitive to these universal color patterns, and they can be used to launch rapid and universal attacks under physical scenes. Therefore, we conduct a universal attack where we use these color patterns and capture 100 photos each with 1) \{$+30^\circ$ light source angle, $0^\circ$ shooting angle\}, and 2) \{$0^\circ$ light source angle, $+30^\circ$ shooting angle\}. Table~\ref{tab:universal_physical} reports the performance of the universal physical attack using three universal color patterns. In the worst case, the AdvColor photos can paralyze the whole face recognition system at an OASR of 87\%. This provides the possibility of physical attacks where the attacker deploys an illumination device to produce illuminations with a fixed universal color near the face recognition system. Due to the reusability of the universal colors obtained by AdvColor, the attacker can insistently use the same universal colors to cause misclassification. For defenders, such a universal attack can reveal multiple universal RGB values to which the pipelines are sensitive, which can help model developers realize the vulnerability of the models they trained or managed, and take timely measures and precautions to avoid such color loopholes. 

The purpose of universal attacks is to reveal the color vulnerability of DNN models. Therefore, the universality among different models might exist, but this is out of the scope of this paper. Despite this, We could still provide two pieces of minor evidence. The first one is the results of universal attack on CelebA-Spoof in Section~\ref{discuss:more_dataset}. The second one is the results of AdvCF~\cite{hu2022colorfilm}. It uses color films to attack ResNet50 on ImageNet, and Figure 8 in their paper also indicates that some colors provide a high fooling rate among different images. Coincidentally, the color with the highest fooling rate (the least classification accuracy of 49.27\%) is purple (255,0,255), which is within the top-1 universal color pattern in our attack - we also find the purple pattern (corresponding to Color ID 1 in Table~\ref{tab:universal_physical}) leads to the highest fooling rate.

\begin{table}[t]  
\centering
\caption{Results of universal physical attacks.}
\vspace{-1mm}
\label{tab:universal_physical}
\resizebox{0.8\linewidth}{!}{
\begin{tabular}{cccccccc}
\toprule
\multirow{2}{*}{\textbf{Color ID}} & \multicolumn{3}{c}{\textbf{light source angle $ = 30^\circ$}} & \multicolumn{3}{c}{\textbf{shooting angle $ = 30^\circ$}} \\
\cmidrule(r){2-4}\cmidrule(r){5-7}
& \textbf{FR}$\uparrow$ & \textbf{AQS}$\uparrow$ & \textbf{OASR}$\uparrow$ & \textbf{FR}$\uparrow$ & \textbf{AQS}$\uparrow$ & \textbf{OASR}$\uparrow$ \\
\midrule
1 & 95\% & 0.51 & 87\% & 60\% & 0.44 & 42\% \\
2 & 83\% & 0.52 & 64\% & 55\% & 0.42 & 35\% \\
3 & 77\% & 0.49 & 52\% & 42\% & 0.37 & 24\% \\
\bottomrule
\end{tabular}
}
\vspace{-3mm}
\end{table}

\subsection{Experimental Results on Countermeasures}
To test the robustness of AdvColor against adversarial defenses, we select ten defensive methods to evaluate adversarial images generated by AdvColor. These methods include proactive defenses (\ie preprocessing, denoising, robustness enhancement) and reactive defenses (\ie adversarial attack detection). Since most defensive methods are related to the overall performance of the classification model, we add bonafide images to the training process of these defenses and take into account the misclassification of both bonafide and spoofed images at the same time. All defense models are based on CDCN++~\cite{yu2020cdcn} and are trained from scratch. Table~\ref{tab:defense_performance} reports the performance of the ten defense approaches. We train the models of RS and TRADES on clean images. The other models are trained on a mix of both colorized and clean images. For proactive defenses, we report the AUC on clean images (cAUC) and on both clean images and colorized images with five randomly chosen colors (ccAUC). The dMR (defensive misclassification rate) is also reported to show how many colorized images are misclassified by the defensive method. Since adversarial detectors are binary classification models that distinguish adversarial images from clean images, we report the detector AUC (dAUC) for detection methods.

Table~\ref{tab:defense_performance} shows that most defenses can achieve a high cAUC, but their performance on ccAUC is quite poor. This implies that the performance of existing defenses against unrestricted perturbations such as AdvColor is limited. We would like to point out that the defense performance of LID~\cite{ma2018lid} is better than that of others, but still does not close the security gap brought by AdvColor. A possible explanation would be that LID can effectively differentiate inputs whose statistics are distinct from those of normal training samples. Since AdvColor drastically shifts the pixel distribution of the input, the discrepancy between the LID statistics of the adversarial image and the clean image becomes relatively significant. 

We also provide an adaptive defense, called ColorAT, based on adversarial training. In each training step, we find a universal color perturbation maximizing the classification error of the current minibatch of data and then update the model to minimize the error. The optimization problem is formulated as a minimax game as follows:
{\small
\begin{equation}\label{eq:adversarial_training1}
\mathop {\min }\limits_\theta  {\mathbb{E}_{{x_b}\sim X}}\left[ {\mathop {\max }\limits_{{\alpha _f}} {\mathbb{E}_{x\sim {x_b}}}\mathds{1}\left[ {f\left( {\theta ;t_r\left( \Theta \left( {x,{\alpha _f}} \right) \right)} \right) \ne {y_0}} \right]} \right],
\end{equation}
}%
where $\theta $ denotes the model parameters, $x_b$ denotes a minibatch of data from the dataset $X$, $y_0$ denotes the clean label for $x$, $t_r$ denotes a random transformation different from the transformation set used in the attack. The defense performance of ColorAT is also reported in Table~\ref{tab:defense_performance}. The trained model maintains
a cAUC of 0.86 on clean images. When attacking the trained model using AdvColor, experimental results (FR: 85\%, AQ: $>$3.3w, OASR: 77\%) show that ColorAT demonstrates a certain defense capability,
since more queries are needed and both FR and OASR fall.

\begin{table}[t]  
\centering
\caption{Defense performance against AdvColor.}
\vspace{-1mm}
\label{tab:defense_performance}
\resizebox{0.85\linewidth}{!}{
\begin{tabular}{ccccccccccccccccc}
\toprule
\textbf{Type} & \textbf{Method} & \textbf{cAUC$\uparrow$} & \textbf{ccAUC$\uparrow$} & \textbf{dMR$\downarrow$} \\
\midrule
\multirow{1}{*}{Preprocessing}
 & Mustafa \etal~\cite{mustafa2019image} & 0.72 & 0.51 & 0.58 \\
\midrule
\multirow{2}{*}{Denoising}
 & Xie \etal~\cite{xie2019feature} & 0.87 & 0.54 & 0.44 \\
 & Comdefend~\cite{jia2019comdefend} & 0.76 & 0.50 & 0.56 \\
\midrule
\multirow{6}{*}{\makecell[c]{Robustness\\Enhancement}}
 & Gong \etal~\cite{gong2017adversarial} & 0.97 & 0.62 & 0.45 \\
 & vanilla AT~\cite{madry2017towards} & 0.72 & 0.51 & 0.51 \\
 & Ensemble AT~\cite{Tramèr2018ensemble} & 0.75 & 0.56 & 0.49 \\
 & RS~\cite{jeremy2019certified} & 0.98 & 0.60 & 0.40 \\ 
 & TRADES~\cite{zhang2019trades} & 0.87 & 0.50 & 0.47 \\
\midrule
Adaptive Defense & ColorAT & 0.86 & 0.79 & 0.19 \\
\bottomrule
\textbf{Type} & \textbf{Method} & \textbf{dAUC$\uparrow$} \\
\midrule
\multirow{2}{*}{\makecell[c]{Adversarial Detector}}
 & DeepDetector~\cite{liang2018deepdetector} & 0.52 \\
 & LID~\cite{ma2018lid} & 0.71 \\
\bottomrule
\end{tabular}
}
\end{table}

\section{Human-Perspective Experience}\label{sec:summary_of_experience}
The excellent attack performance and ability to circumvent adversarial defenses prompt us to think about why such a simple attack can pose a great security threat. Although the academic community has been committed to proposing novel, technically complex attacks to achieve imperceptible perturbations, the industry might possess different opinions towards the attacks. This motivates us to conduct a comprehensive survey on threat measurement from the viewpoints of industry professionals to close the cognitive gap. There are three key research questions (RQs) we want to answer in this section.
\begin{itemize}
    \item[1)] \emph{Do professionals think the risk increases with the imperceptibility of perturbations?}
    \item[2)] \emph{What is a more instructive measurement for the threat level of attacks?}
    \item[3)] \emph{What is the industry's opinion on adversarial attacks?}
\end{itemize}
Note that the \textit{risk} is defined as the likelihood of a particular \textit{threat} exploiting a particular \textit{vulnerability} of face recognition systems, resulting in severe damages. 

\subsection{Survey Details}
We designed an online survey that included multiple choice questions and subjective questions encircling the risk of various adversarial attacks. We recruited 40 anonymous subjects who are over 18 years old and work in the industry to answer the survey posted on an online platform called Amazon Mechanical Turk (AMT)~\cite{amazon_turk}. The Human Research Ethics Committee of the lead author’s affiliation determined that the study was exempted from further human testing and we followed best practices for conducting ethical research in human subjects. Please refer to Appendix~\ref{append:survey} for more details about our survey (\eg demographics and payment).

\subsection{Intuition Regarding Imperceptibility}
To obtain the subject's intuitive view on the risk of imperceptibility of perturbations in real-world scenarios, we first provided representative examples of restricted adversarial attacks and various kinds of unrestricted attacks mentioned in Section~\ref{subsec:taxonomy} (AdvColor also considered), followed by introductory descriptions and the attack performance. The examples were secured from their original paper. The subjects were then asked which type of attack could be more harmful, as well as easier to realize/implement (\eg by leveraging simple equipment) in real-world scenarios. The harmfulness indicates the consequence of face recognition systems being exploited by attackers who carry out malicious attacks and access sensitive personal information of victims. The ease of implementation measures the practicality of the attackers launching attacks in a cheaper and simpler way. We use these two factors to showcase the risk. Afterwards, we asked the subjects to rank the perceptibility of these attacks according to their cognition and intuition, based on a Likert scale~\cite{likert1932technique} from 1 to 10. A higher score indicates that the perturbations are more obvious.

\begin{figure}[t]
\begin{center}
  \includegraphics[width=0.9\linewidth]{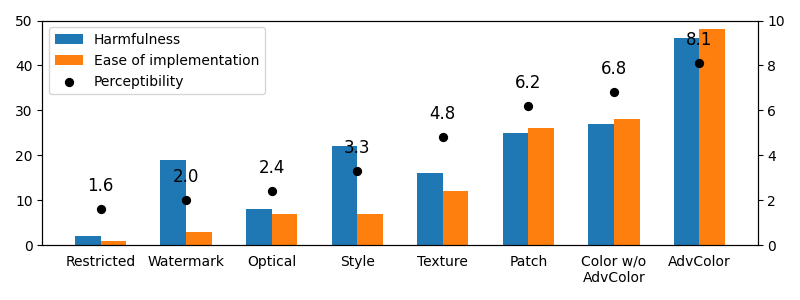}
\end{center}
\vspace{-4mm}
  \caption{Results of harmfulness, ease of implementation and perceptibility.}
\label{fig:harmfulness}
\vspace{-2mm}
\end{figure}

Amazingly, we find that 78\% (98\%) of subjects considered unrestricted attacks more harmful (easier to implement) in real-world scenarios. Results in Figure~\ref{fig:harmfulness} indicate that AdvColor, color, and patch attacks are the top three attacks when considering harmfulness (also the most perceptible), while restricted attacks (obtaining the least perceptibility score) pose the least harm. AdvColor is recognized by 96\% of subjects as the least challenging attack to implement in the real world. The subjects provided an explanation that one only needs to change environment illumination through a smart bulb despite the apparent lighting anomalies that may exist, while style and optical attacks require attackers to wait for the background until certain natural conditions are met. Generally, there is a negative correlation between imperceptibility and real-world risk (including harmfulness and ease of implementation) for industry professionals. Unlike the mainstream research perspective in academia (almost all academic research prioritizes imperceptibility), industry employees had overwhelming concerns about perceptible but effective attacks such as AdvColor. 

\vspace{2mm}
\begin{mdframed}[backgroundcolor=white!10,rightline=true,leftline=true,topline=true,bottomline=true,roundcorner=2mm,everyline=true,nobreak=false]  
\noindent\textbf{Finding summary for RQ 1:}
\begin{itemize} [leftmargin=*]
\item Physical risk and imperceptibility of perturbations are largely negatively correlated, which is inconsistent with the mainstream opinion in academia.
\end{itemize} 
\end{mdframed}
\vspace{-2mm}

\subsection{Instructive Measurements for Threats}
After the subjects answered a few general intuitive questions, we performed a specific evaluation of the threat. We first asked the subjects what the main factors are when considering the threat of attacks. We find that all the subjects agreed that the cost of attack equipment should be very low. Over 90\% of the subjects regarded simple equipment and the ability to realize short-term mass production of adversarial samples as factors of secondary importance. The low threshold for equipment usage and knowledge also received support from 70\% of the subjects, and the simplicity and comprehensibility of the technical principles obtained 24\% of the votes. Statistical results indicate that industry employees prioritize attack costs and mass production capabilities, with little concern for the complexity of technology and theory when maintaining imperceptibility. However, the academia often ignores the former and only focuses on the latter when studying adversarial samples.

We thus summarize the three main factors according to their opinions: 1) \textbf{Weak professionalism (WP)}: Attackers do not need to master relevant professional knowledge, such as deep learning, image recognition, \etc. The threshold for attacks is low, and attacks that ordinary people can launch consciously or unconsciously are more dangerous.
2) \textbf{Ease of mastery (EM)}: Through self-learning or being taught by others, attackers can easily implement and quickly master this form of attack.
3) \textbf{Ease of operation (EO)}: Attack equipment is inexpensive and easy to deploy, and adversarial samples can be produced in large quantities in a short period of time. 
The subjects were then asked to rank the importance of several main factors that we summarize and rate them for different types of attacks in real-world scenarios based on a Likert scale from 1 to 5. A higher score indicates greater threat.

\begin{figure}[t]
\begin{center}
  \includegraphics[width=0.9\linewidth]{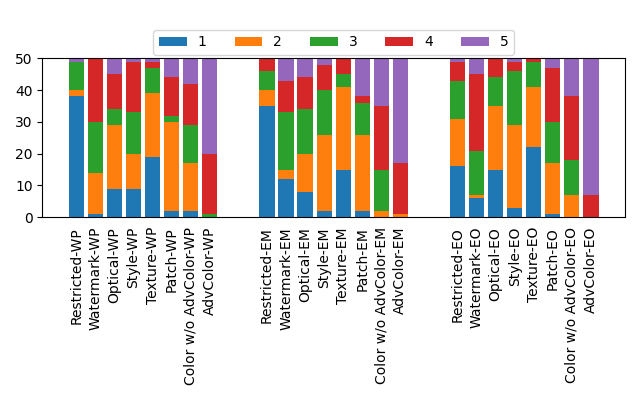}
\end{center}
\vspace{-4mm}
  \caption{Factor rating results.}
\label{fig:factor_rating}
\vspace{-3mm}
\end{figure}

The importance ranking results suggest that implementation simplicity is the most crucial factor since most subjects (over 40\%) voted for the following sequence: Ease of operation $>$ Ease of mastery $>$ Weak professionalism. Figure~\ref{fig:factor_rating} shows the subject's rating results. AdvColor received the highest number of votes for all three factors, indicating that AdvColor is the most threatening attack. This is reflected in the low knowledge threshold required for attackers, simple methods that can be quickly mastered, low equipment costs, and the ability to produce in large quantities. Color attacks (AdvColor excluded) ranked second, since ordinary people had a better understanding of just changing objects of different colors (\eg cAdv~\cite{bhattad2020unrestricted}). Patch attacks also attracted relatively high scores, but they were still much lower than AdvColor because the color and texture of patches are irregular, requiring fine optimization, and the location and size of patches are difficult to determine, making it difficult for batch production. However, restricted attacks obtained the lowest score due to their complicated methodology, complex deployment, and high attack costs.

\vspace{2mm}
\begin{mdframed}[backgroundcolor=white!10,rightline=true,leftline=true,topline=true,bottomline=true,roundcorner=2mm,everyline=true,nobreak=false]  
\noindent\textbf{Finding summary for RQ 2:}
\begin{itemize} [leftmargin=*]
    \item Industry focuses more on attack equipment cost, equipment deployment and mass production capability when considering the threat level of adversarial attacks. To summarize, we propose three instructive measurements, \textit{weak professionalism}, \textit{ease of mastery}, and \textit{ease of operation} to replace imperceptibility.
\end{itemize} 
\end{mdframed}
\vspace{-3mm}

\subsection{Industry’s Perspectives}
Finally, we asked about some personal perspectives of the subjects towards adversarial attacks and defenses, as well as the security issues they met and potential consequences.

We find that all the subjects considered that defending against perceptible attacks (\eg AdvColor) necessitates, the reasons for which were twofold. First, these attacks have a wide range of applications and can be implemented in different scenarios. 50\% (mostly algorithm/testing engineers) of the subjects claimed that some customers have already reported issues of perceptible attacks. However, the security control of data links in major software/Internet systems nowadays is relatively mature, which undoubtedly greatly increases the difficulty of restricted attacks. Secondly, perceptible attacks may have a more serious impact on the security and availability of the system. Specific unrestricted attacks may lead to complete system failure or serious vulnerabilities, causing significant reputation and financial losses to the organization. On the contrary, the impact of restricted attacks may be more limited and can only be successfully implemented under specific conditions or environments. Therefore, prioritizing defense against perceptible attacks can reduce the potential risks faced by the system, since these attacks typically have a broader potential threat scale, severer social impacts, and simpler defense requirements. More than half of the subjects (including all the managers) also expressed that the harm of restricted attacks should not be underestimated; they believed that both restricted attacks and unrestricted attacks should be considered in business products to ensure that the system has comprehensive security in the face of various attack threats. One general manager pointed out that restricted attacks are non-communicable, while unrestricted attacks, whose attackers are mostly end individual users, are easy to spread, difficult to trace and take timely defensive measures, and can pose great harm to public opinion. Both attacks need guarding, but unrestricted attacks engender more harm, especially for the financial industry.

When asked to imagine if AdvColor or other perceptible attacks that ordinary people can intentionally or unintentionally launch occurred on the products they develop or handle, almost all subjects considered such a color hazard as a severe security loophole -- it poses an Advanced Persistent Threat (APT) to web and mobile systems. We outline the main aftermaths mentioned by the subjects. 
1) \textbf{Business damage}: Perceptible attacks may cause serious impact or even paralysis on the business, including financial losses, legal liability, and reputation damage. For products directly related to financial rights (\eg payment verification, loan risk control, equity issuance), being breached will directly lead to the abandonment of the entire product.
2) \textbf{Customer trust damage}: Companies will face doubts and challenges from both customers and society about the level of products and technology due to security vulnerabilities or system failures in products or services. Customers will lose confidence and turn to secure competitors/alternative.
3) \textbf{Regulatory compliance issues}: If perceptible attacks result in data leakage or violation of compliance requirements, companies may face legal proceedings, fines, and reputation damage. Additionally, if the safety of the product cannot meet specific industry standards or regulatory requirements, it may face risks, such as sanctions or sales bans from regulatory agencies. 
4) \textbf{Personal privacy leakage}: Perceptible attacks may increase the risk of personal privacy, identity theft, and other illegal activities, especially in face-recognition-related systems where attackers can use other's face to cause mismatching and access sensitive personal information.
5) \textbf{Potential product security}: The success of perceptible attacks may expose product security issues and may reveal potential breaches and vulnerabilities for other similar types of attack, the threat of which will directly affect user satisfaction and long-term sustainable development of the enterprise. We leave more details to Appendix~\ref{append:survey_details}.

\vspace{2mm}
\begin{mdframed}[backgroundcolor=white!10,rightline=true,leftline=true,topline=true,bottomline=true,roundcorner=2mm,everyline=true,nobreak=false]  
\noindent\textbf{Finding summary for RQ 3:}
\begin{itemize} [leftmargin=*]
\item Industrial insiders generally believe that perceptible attacks have seriously threatened the security of products, and may invite negative impacts on business, customers, privacy leakage, and other aspects. Business professionals are appealing for more research-oriented focus on perceptible but effective attacks and their mitigation.
\end{itemize} 
\end{mdframed}
\vspace{-3mm}

\section{Discussion}
\subsection{Why We Choose These Techniques}\label{subsec:pso}
The main purpose of our work is to expose the vulnerability of face recognition pipelines to this easy-to-launch but critical illumination attack, rather than the technical complexity. We use PSO because there are only three optimization variables $\alpha_f  = {\left[ {{r_f},{g_f},{b_f}} \right]^T}$, and PSO is a relatively efficient optimization method in this case due to its search efficiency and flexibility. Though widely used, EOT can help us make images adaptable to physical scenarios, and we additionally consider several transformations which are exclusively required in an illumination-based attack. On the other hand, switching to other sophisticated and novel methodological techniques may only bring marginal improvements in the optimization performance but may significantly increase the computational/query cost. Compared to the slight improvement in efficiency and performance, we underscore the simplicity and reusability of the proposed attack.

Color transformation may also not be a brand-new technique in image attacks, but there are still discrepancies between our work and several existing works. SAE~\cite{hosseini2018semantic} converts the RGB space to HSV space and adds random perturbations to H and S channels. However, the attack performance depends on the randomness, which makes the trial number uncertain (the same for ColorFool~\cite{shamsabadi2020colorfool}). Despite over 90\% attack success rate after 1,000 trials on CIFAR10, the query budget would remain high and unstable when attacking larger models on larger datasets. cAdv~\cite{bhattad2020unrestricted} introduces color transformation by utilizing a pretrained colorization model and jointly varies both hints and masks to craft adversarial perturbations. cAdv is a white-box attack in general which underestimates the ability of the adversary. cAdv also uses K-Means to segment several regions to determine which region should be added with color transformation, which leads to high deployment difficulty and cost in the real world. To recap, both SAE and cAdv are not applicable in physical scenarios, and the attack setting is much easier than that of AdvColor, which is a hard-label black-box physical attack.

\subsection{Experiments on More Datasets}~\label{discuss:more_dataset}
Our work focuses more on the vulnerability of the face recognition pipeline to this illumination attack, and we aim to reveal and narrow the cognition gap between academia and industry. Therefore, we weaken the experimental section and only use Replay-Attack, which is representative and widely used in this field, as our dataset to validate the effectiveness of our attack. To show the effectiveness of our method on more diversified data, we conduct additional experiments on a larger dataset CelebA-Spoof~\cite{zhang2020celeba}, which contains 625,537 images from 10,177 subjects. We use the AENet proposed in their paper as the target model. We randomly choose 500 images from the test set and launch the attack. We find that our attack still achieves an FR of 81.2\%, with an AQ of 235.1. We examine the samples that did not succeed in the attack and find that almost all of their spoofing types were masks, and many images have facial tilt angles that are too large during shooting (\eg looking down at the ground at an angle of almost 90 degrees). However, in general, facial detectors cannot detect faces at this time. Therefore, we believe that the presence of dirty data affects the performance of attacks. On the other hand, when attacking in physical scenarios, attackers often only choose to use photos (in types of prints or replays, which are the main types in ReplayAttack) of the victim for the attack, rather than using masks. The cost of making masks for victims is very high, which goes against our pursuit of low-cost and accessible attacks. However, various types of masks account for approximately 40\% of the spoofing images in CelebA-Spoof. Therefore, we believe that although our fooling rate has not reached nearly 100\%, the demand for attacks against masks in physical scenarios is very low.

Due to the simplicity and accessibility of prints and replays, we randomly choose 10,000 spoofing images from CelebA-Spoof which are categorized to prints and replays to conduct the attack. The experimental results show that the fooling rate reaches 98.14\% at this time. We also use the universal color selected from Replay-Attack to test the transferability of universal colors between different models. We surprisingly find that the fooling rate is 31.6\% for Color ID 1 and 10.5\% for Color ID 2 (these two colors refer to those in Table~\ref{tab:universal_physical} for Replay-Attack), which indicates that even models trained on diversified data may still be sensitive to some specific color illumination.

\subsection{Comparisons with Other Color Attacks}~\label{discuss:other_color_attacks}
There are some technical differences between AdvColor and other color-based attacks such as AdvCF~\cite{hu2022colorfilm} and NCF~\cite{yuan2022natural}, which make it difficult to ensure fair comparison.

AdvCF is actually a soft-label white-box attack since it requires the confidence score of the ground-truth label during the optimization. The adversarial images generated by a local model are then fed into other unseen models to achieve transfer-based attacks. The soft-label setting reduces the attack difficulty and the transferability also weakens the black-box setting. Similarly, NCF is also a soft-label transfer-based attack. It first attacks the local substitute model under an outright white-box setting by using the C{\&}W loss~\cite{carlini2017towards} and then tests the attack success rate on some unseen models to verify its transferability. By contrast, AdvColor is a hard-label black-box attack, where the attacker can only access the top-1 label (no score of either the top-1 label or the ground truth label) and no need for a local surrogate model. Overall, it is not fair and scalable to compare AdvColor with these attacks due to different attack settings and attack scenarios.

We also provide detailed analyses between AdvColor and other color-based attacks including AdvCF, NCF, \etc. AdvCF attaches a color film in front of the camera to simulate the color film effect. However, color films add no depth information to the object since AdvCF can be seen as positioning a color filter between the camera and the object in the physical scenarios. In contrast, AdvColor uses a color source to illuminate the object, which provides a three-dimensional illumination effect from the external environment and therefore brings about more verisimilitude. There is also another advantage for attacking face anti-spoofing models. Most anti-spoofing models have the ability to distinguish two-dimensional features (\eg prints, replays) and three-dimensional features (real faces). Therefore, adding more depth information to the prints can increase the possibility of misguiding the anti-spoofing models.

NCF, as well as ColorFool~\cite{shamsabadi2020colorfool}, is a digital attack rather than a physical attack. NCF and ColorFool require fine-grained object segmentation before the attack, and the methodology is rather complicated. Attacks such as cAdv~\cite{bhattad2020unrestricted}, ColorFool and NCF all consider adding different color transformations to different semantic regions, which may be applicable in digital images, but is not suitable to physical scenarios since it is rather difficult or even unrealistic to deploy different objects with various colors. One main improvement of NCF compared to ColorFool is to make the color more realistic in the real world. However, this amelioration is marginal in unattended scenarios such as access control systems, where imperceptibility is no longer a significant requirement.

To conclude, as a hard-label black-box attack, it would be more fair to make comparisons with other hard-label attacks, rather than the existing color attacks. We argue that, as analyzed above, comparing with these attacks is possible, but unnecessary.

\section{Conclusion}
This paper studies the threat assessment of adversarial attacks towards face recognition systems from the industry perspective. We first propose a black-box easy-to-exploit physical attack, AdvColor, by attacking the printed photos of human faces under the adversarial illuminations. AdvColor is effective and efficient, yet the devices required for launching the attack are cheap and accessible. AdvColor not only achieves high fooling rates against the face anti-spoofing model, but also retains the face quality score and keeps the identity in face matching in both simulation and physical scenarios. We also identify that AdvColor can help find several sample-agnostic adversarial colors for crafting large batches of adversarial examples, which allows attackers to place illumination equipment and fix illumination color near the face recognition system for persistent physical attacks. Additionally, as an unrestricted attack, AdvColor can bypass ten advanced adversarial defenses with ease. We otherwise propose an adaptive defense, called ColorAT, which has a certain defensibility against colorized samples. Afterwards, we conduct a human survey on the threat level of different types of attacks. We find that industry professionals believe that imperceptibility is not the most crucial factor in assessing the risk of attack. Instead, they pay more attention to the cost of attack equipment and mass production capabilities even if the perturbations are perceptible. We would like to draw the attention of researchers in both academia and industry to such simple but harmful attacks in order to protect the privacy and security of such responsible AI techniques in the future.

\small
{
\bibliographystyle{IEEEtran}
\bibliography{reference}
}

\clearpage
\appendix

\subsection{Demographics and Payment About the Survey}\label{append:survey}
In order to facilitate a deeper understanding of the subjects' personal situation, we first asked for some basic information, including age, working sector, position, role in the job, \etc. We did not collect unnecessary personal information that was not related to our survey. 

Table~\ref{tab:basic_info} illustrates the results. Clearly, most of the participants were between 26 and 35 years old and worked as algorithm/testing engineers in state-owned or private companies. Therefore, the answers of this majority of working professionals are representative in reflecting the mainstream opinions of contemporary young employees about adversarial attacks and defenses. We also received seven and three questionnaires from project/product managers and (deputy) general managers, respectively. These answers might be significant and unique, since they were regarding problems on a higher, broader, and macro level. It is also noticeable that 44 participants worked on vocational content only, and only 6 subjects worked mainly on business work, which means that they might be slightly exposed to some academic work. We did not recruit subjects working on research only since this paper focuses on revealing the threat of adversarial attacks in the industry. We further inquired about the specific work directions of the subjects, most of which were autonomous driving, facial recognition, virtual reality, financial technology, smart healthcare, \etc. Surprisingly, we found that all subjects claimed that they were familiar or partially familiar with adversarial attacks, which helped us to eliminate possible errors caused by random responses from those without a relevant knowledge background.

In our user study, we only hired participants over 18 years old from USA, China and Australia with an answer approval rate of at least 95\%. The survey was projected to be finished within 30 minutes, thus we paid each participant \$20.00, which was much higher than the average hourly payment (\$11.00) on AMT~\cite{hara2018data}.

\subsection{More Analyses about the Survey}\label{append:survey_details}
\noindent\textbf{Digital attack simplicity.}
From the perspective of digital attacks, we also briefly introduced the core steps required for each attack in digital implementation (\eg gradient descent, reinforcement learning) to the subjects and asked them to rate the attack simplicity, based on a Likert scale~\cite{likert1932technique} from 1 to 5. A higher score indicates that the attack is simpler. 

\begin{figure}[t]
\begin{center}
  \includegraphics[width=0.9\linewidth]{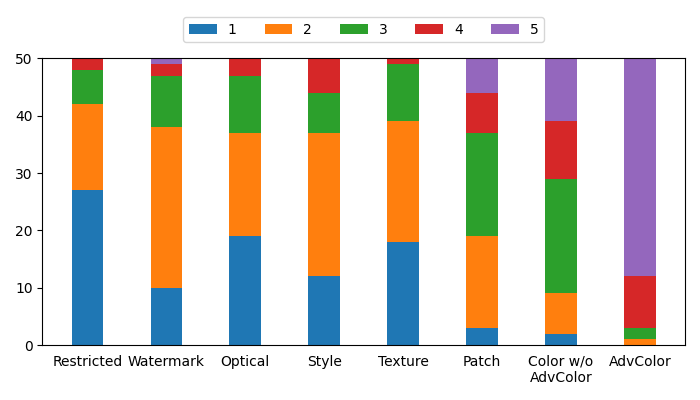}
\end{center}
\vspace{-5mm}
  \caption{Results of the simplicity of digital implementation.}
\label{fig:simplicity}
\end{figure}

Figure~\ref{fig:simplicity} shows the rating of simplicity. It is obvious that AdvColor enjoys the highest simplicity score, with 76\% and 18\% of subjects rating 5 and 4 respectively. We believe that this phenomenon is caused by the fewer optimization parameters required during the attack process (\eg only three channels of the color filter in AdvColor). In contrast, other attacks require either more optimization parameters or a large amount of computational cost during the optimization or training stage, leading to lower rating scores. The results of the digital simplicity rating in Figure~\ref{fig:simplicity} and the ease of implementation in Figure~\ref{fig:harmfulness} reflect that the proposed AdvColor is simple in both digital and physical scenarios.

\noindent\textbf{Business threat.}
Regarding the threat to business products posed by adversarial samples produced in large quantities using some simple technologies by some black companies, the result showed that 24\% (including three (deputy) general managers) of the subjects thought the impact is significant and it is necessary to develop defense strategies as soon as possible to address this challenge. 50\% (mostly algorithm/testing engineers) of the subjects claimed that some customers have already reported issues of adversarial attacks. Only 26\% of the subjects had not observed similar problems but they were not optimistic about the future. Totally unexpectedly, no one deemed that there is almost no impact, even though we set this option.

\noindent\textbf{Viewpoints on AdvColor.}
We briefly introduced our proposed attack AdvColor, and asked for their opinions on this form of attack, especially in vocational scenarios. Most subjects shared the viewpoint that AdvColor is featured with simplicity, low cost, and low threshold. Moreover, subjects came to an agreement that AdvColor poses an Advanced Persistent Threat (APT), which may lead to a decrease in the reliability and accuracy of the system. Once the light source or smart light bulb is placed and begins to emit specific colored light, it can continuously affect the performance of the access control system or facial recognition system. Some subjects also expressed concerns about the security issues and public opinion caused by this simple but powerful attack on business products.

\begin{table}[tb]  
\centering
\caption{Demographics of participants.}\label{tab:basic_info}
\resizebox{0.98\linewidth}{!}{
\begin{tabular}{cccccccc}
\toprule
\multicolumn{2}{c}{\textbf{Age}} & \multicolumn{2}{c}{\textbf{Sector}} 
& \multicolumn{2}{c}{\textbf{Position}} & \multicolumn{2}{c}{\textbf{Role}} \\
\midrule
18-25 & 3 & Multinational & 5 & Algorithm/Testing Engineer & 39 & Vocational only & 44  \\
26-35 & 32 & State & 18 & Project/Product Manager & 7 & Vocational mainly & 6  \\
36-45 & 11 & Private & 24 & (Deputy) General Manager & 3 & Research only & 0 \\
$>$ 45 & 4 & Other & 3 & Other & 1 &  &  \\
\bottomrule
\end{tabular}
}
\end{table}

\noindent\textbf{Potential solutions to defend AdvColor.}
From the perspective of industry professionals, possible solutions to perceptible attacks such as AdvColor were mainly focused on improving data preparation (\eg data augmentation), model structure (\eg adding a sub-module to the model and decoupling the color components), and loss function (\eg assign different weights to unbalanced data). A project manager also suggested proposing a quantitative formula to measure color sensitivity. Meanwhile, a few subjects also worried that data augmentation could affect the judgment results in the already objectively existing colored lighting environment.

\noindent\textbf{Potential consequences of attacks such as AdvColor.}
Finally, we asked the subjects to imagine a vocational scenario where the model used by the customers (with sufficient accuracy on the test set) is found to have security issues, \eg shining purple light on all faces of an access control system can cause misclassification in most cases. we asked the subjects about the aftermath or harm if perceptible attacks such as AdvColor occurred on the products they develop or handle. We summarize the responses of the subjects from the following five perspectives. 

    1) \textbf{Business damage}: Unrestricted attacks may cause serious impact or even paralysis for the business. For example, in facial recognition systems, if attackers can use AdvColor to bypass recognition mechanisms, they may be able to enter restricted areas or perform unauthorized operations, thereby endangering security and confidentiality. This may lead to business consequences such as financial losses, legal liability, and reputation damage. For products directly related to financial rights (\eg payment verification, loan risk control, equity issuance), being breached will directly cause economic losses to customers, leading to abandonment of the entire product. 
    
    2) \textbf{Customer trust damage}: Security vulnerabilities or system failures may cause serious damage to customer trust in products or services, as well as doubts and challenges about the level of products and technology from the society. If customers' personal information or sensitive data is attacked, they may lose confidence in their products or services and choose to turn to competitors or find more secure alternatives. This will have a negative impact on the customer relationship and market competitiveness of the enterprise. 
    
    3) \textbf{Regulatory compliance issues}: Some industries and regions may have strict data protection and privacy regulations. If perceptible attacks, such as AdvColor, result in data leakage or violation of compliance requirements, companies may face legal proceedings, fines, and reputation damage. In addition, if the safety of the product cannot meet specific industry standards or regulatory requirements, it may face risks such as sanctions or sales bans from regulatory agencies. 
    
    4) \textbf{Personal privacy leakage}: AdvColor may increase the risk of personal privacy leakage. For example, in facial recognition systems, attackers may use AdvColor to mislead the system, identify incorrect faces, and access sensitive personal information, which may lead to personal privacy leakage, identity theft, and other illegal activities. 
    
    5) \textbf{Potential product security}: AdvColor exposes product security issues and may reveal potential breaches and vulnerabilities. This may make it easier for attackers to engage in other types of attacks, and may even exploit known vulnerabilities to invade the system, tamper with data, or engage in illegal operations. The threat to product safety will directly affect the reputation, user satisfaction, and long-term sustainable development of the enterprise. 

In summary, perceptible attacks such as AdvColor may bring serious consequences and harm to businesses, customers, and product security. Therefore, it is crucial to ensure the security and robustness of the products, respond to and fix potential security vulnerabilities in a timely manner, and strengthen security testing and defense measures. In addition, establishing a close monitoring mechanism and response plan, as well as complying with applicable laws and regulations, is also crucial to mitigate potential risks and protect the interests of enterprises and customers.

\subsection{Justification for Anti-Spoofing}\label{append:justification}
In this section, we summarize and emphasize the crucial role of anti-spoofing components in the face recognition pipeline, especially the relationship between face recognition systems and face anti-spoofing.

Attacks against face recognition systems usually fall into two categories: digital deception~\cite{tolosana2020deepfakes_and_beyond} and physical presentation attacks~\cite{liu2021cross-ethnicity}. The former deceives the face system through visual operations that are difficult to observe in the digital field, which is similar to the adversarial samples in the image field. The latter usually misleads the real-world face recognition system by presenting a face on the physical medium in front of the imaging sensor (\eg prints, replays, masks). The images generated by physical presentation attacks are also called spoofed images, which is opposite to the concept of bonafide images (\ie living objects of real faces).

Spoofed images usually include impersonation and obfuscation. Impersonation imitates the victim's facial features by means of paper printing, image replay or other media such as masks and mannequins, thus cheating the face recognition system. Obfuscation hides the identity of the attacker by making up, wearing glasses, wigs, etc. In addition, spoofed images can also be divided into 2D attacks~\cite{ramachandra2017presentation} and 3D attacks~\cite{jia20203dmask}. 2D attacks collect spoofed images by taking pictures on paper and electronic screens with mobile phones or visual sensors. 3D attacks use mature 3D printing technology to print a 3D head mold, which can be made of resin, plaster, latex and other materials. The printed head mold is very lifelike, and it is difficult for the human eye to distinguish. There are also a few attacks against some areas of the face, such as sticking the printed nose to the real face, wearing glasses with special textures, and sticking tattoos on the face.

In order to prevent spoofed images, the anti-spoofing classification model, together with a quality score assessment, is usually added before face matching to distinguish whether the input faces are bonafide or spoofing.  Often, there are two specific integration methods: parallel integration~\cite{de2013can} and cascading integration~\cite{li2018face}. Parallel integration comprehensively judges whether to reject face matching according to the output results of anti-spoofing classification and face recognition respectively. Cascading integration requires the input image to pass through the anti-spoofing classification model first. Only the faces recognized as bonafide will be sent to the face recognition system for face matching, and the spoofed images will be discarded in the anti-spoofing classification stage. 

As mentioned before, anti-spoofing plays the most important role in the face recognition pipeline. Obviously, from the attacker’s view, whether the face recognition pipeline is a parallel or cascading structure, if the attacker cannot misguide the face anti-spoofing detection model from spoofing to bonafide, even if he can attack face quality evaluation and face matching, it will be of no use. Especially in the cascading structure, the attacker cannot even access the subsequent components if the input face is already rejected by the anti-spoofing model. Methodologically, face quality assessment only needs simple face feature extraction, and face ID matching is mainly depending on the calculated cosine similarity between the input face and the known IDs. However, face anti-spoofing models are more complicated since they have to extract more features to get the information of the input image, \eg in order to distinguish 3D masks from real faces, a larger and deeper network is required to obtain high-level depth information and sometimes a fine-grained detection is required to check if there are some splicing traces in the face region. To conclude, anti-spoofing is the most important part of the face recognition pipeline, and also the most difficult component to launch the attack.

\end{document}